\documentclass[acmsmall]{acmart}

\AtBeginDocument{%
  }

\setcopyright{acmlicensed}
\copyrightyear{2026}
\acmYear{2026}
\acmDOI{XXXXXXX.XXXXXXX}

\acmJournal{TKDD}
\acmVolume{XX}          
\acmNumber{XX}          
\acmArticle{XX}        
\acmMonth{6}           
\acmYear{2026}

\usepackage{microtype}
\usepackage{graphicx}
\usepackage{subfigure}
\usepackage{booktabs} 

\usepackage{mathtools}

\usepackage{amssymb}
\usepackage{amsthm}
\usepackage{color}

\DeclareMathOperator*{\argmin}{arg\,min}

\newcommand{\cut}[1]{{}}
\newcommand{\xr}[1]{{\color{red}(#1 -- XR)}} 
\newcommand{\zc}[1]{{\color{blue}#1}} 

\newcommand{\tA}{{\tilde{\vA}}}

\newcommand{\tin}{{\text{in}}}
\newcommand{\tout}{{\text{out}}}
\newcommand{\tsym}{{\text{sym}}}

\newcommand{\vA}{{\mathbf{A}}}

\newcommand{\vD}{{\mathbf{D}}}

\newcommand{\vF}{{\mathbf{F}}}

\newcommand{\vI}{{\mathbf{I}}}

\newcommand{\vM}{{\mathbf{M}}}

\newcommand{\vP}{{\mathbf{P}}}

\newcommand{\vS}{{\mathbf{S}}}

\newcommand{\vX}{{\mathbf{X}}}
\newcommand{\vY}{{\mathbf{Y}}}

\newcommand{\cT}{{\mathcal{T}}}

\newcommand{\cV}{{\mathcal{V}}}

\newcommand{\RR}{\mathbb{R}}

\newcommand{\vone}{{\mathbf{1}}}

\makeatletter
\let\@@span\span
\def\sp@n{\@@span\omit\advance\@multicnt\m@ne}
\makeatother

\newcommand{\bc}{\begin{center}}
\newcommand{\ec}{\end{center}}

\newcommand{\bdm}{\begin{displaymath}}
\newcommand{\edm}{\end{displaymath}}

\newcommand{\beq}{\begin{equation}}
\newcommand{\eeq}{\end{equation}}

\newcommand{\bfl}{\begin{flushleft}}
\newcommand{\efl}{\end{flushleft}}

\newcommand{\bt}{\begin{tabbing}}
\newcommand{\et}{\end{tabbing}}

\newcommand{\beqn}{\begin{align}}
\newcommand{\eeqn}{\end{align}}

\newcommand{\beqs}{\begin{align*}} 
\newcommand{\eeqs}{\end{align*}}  

\usepackage{hyperref}

\usepackage{amsmath}

\usepackage{amssymb}

\usepackage{mathtools}
\usepackage{amsthm}
\usepackage{subcaption} 
\usepackage{wrapfig}
\usepackage{multirow}
\usepackage{makecell}
\usepackage{subfigure}
\usepackage{enumitem}
\usepackage[capitalize,noabbrev]{cleveref}
\usepackage{algorithm} 
\usepackage{algpseudocode} 

\usepackage[textsize=footnotesize]{todonotes}

\theoremstyle{plain}
\newtheorem{theorem}{Theorem}[section]

\theoremstyle{definition}
\newtheorem{definition}[theorem]{Definition}

\theoremstyle{remark}
\newtheorem{remark}[theorem]{Remark}

\begin{document}

\title{Harnessing Trust in Directed Graphs: Redefining Robustness of Graph Learning}

\author{Zhichao Hou}
\affiliation{%
  \institution{North Carolina State University}
  \city{Raleigh}
  \country{USA}}
\email{zhou4@ncsu.edu}

\author{Xitong Zhang}
\affiliation{%
  \institution{Michigan State University}
  \country{USA}}
\email{zhangxit@msu.edu}

\author{Wei Wang}
\affiliation{%
  \institution{Meta Inc.}
  \country{USA}}
\email{weiwangmsu@meta.com}

\author{Charu C. Aggarwal}
\affiliation{%
  \institution{
IBM T.J. Watson Research Center}
  \country{USA}}
\email{charu@us.ibm.com}

\author{Xiaorui Liu}
\authornote{Corresponding author.}
\affiliation{%
  \institution{North Carolina State University}
  \city{Raleigh}
  \country{USA}}
\email{xliu96@ncsu.edu}

\begin{abstract}

Existing research on robust Graph Neural Networks (GNNs) focuses predominantly on undirected graphs, overlooking the trustworthiness inherent in the link directions of directed graphs. In this work, we analyze the limitations of current approaches from both the attack and defense perspectives and redefine the robustness of GNNs in directed graphs. From the attack side, we introduce a more realistic directed graph attack setting that addresses the shortcomings of existing attacks. From the defense side, we propose a simple yet effective message-passing framework that serves as a plug-in layer to strengthen GNN robustness while avoiding a false sense of robustness. Our findings show that the trust signals encoded in directed graphs can be harnessed to substantially improve the robustness of GNNs. When combined with existing defense strategies, the framework achieves state-of-the-art robust performance alongside strong clean accuracy against both transfer and adaptive attacks. These results point to a novel and promising direction for strengthening the robustness of graph deep learning.

\end{abstract}


\received{4 February 2025}
\received[revised]{4 January 2026}
\received[accepted]{1 June 2026}

\maketitle

\section{Introduction}
\label{sec:intro}



Graph neural networks (GNNs) have emerged to be a promising approach for learning 
representations from graph data, owing to their ability to capture node features and graph 
topology information 
~\citep{ma2020deep, grl_william, yu2024automated, wu2023equivariant}. However, extensive research has revealed that GNNs are vulnerable to adversarial attacks \citep{dai2018adversarial,jin2021adversarial,hou2024adversarial,wu2019adversarial,zugner2018adversarial,zugner2019certifiable}. Even slight perturbations in the graph structure can lead to significant performance deterioration. Despite the existence of numerous defense strategies, their effectiveness has been questioned due to a potential false sense of robustness against transfer attacks~\citep{mujkanovic2022defenses}. In particular, a recent study~\citep{mujkanovic2022defenses} demonstrated that existing robust GNNs are much less robust when facing stronger adaptive attacks. 
In many cases, these models even underperform simple multi-layer perceptrons (MLPs) that disregard graph topology information, indicating the failure of GNNs in the presence of adversarial attacks. 
As existing research fails to deliver satisfying robustness, new strategies are needed to effectively enhance the robustness of GNNs. 

\begin{wrapfigure}{r}{0.45\textwidth}
 \centering
 \vspace{-0.1in}
 \includegraphics[width=0.43\textwidth]{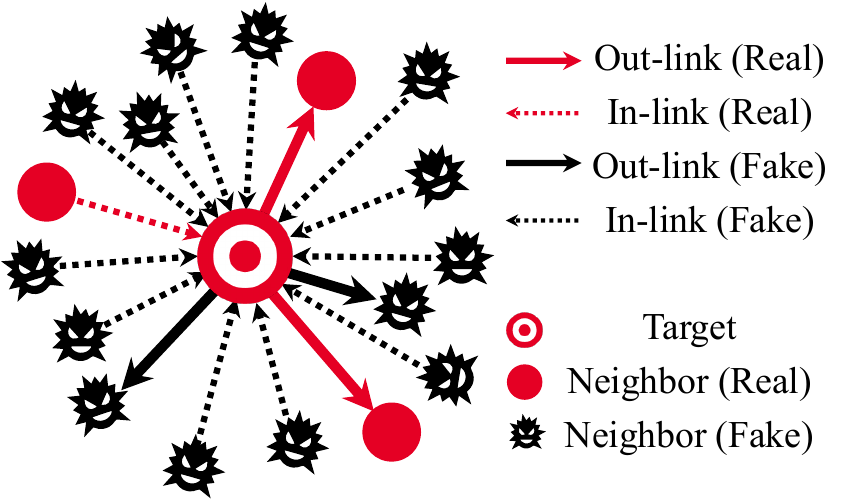}

 \caption{Large-scale link spam. Attackers can cheaply fabricate many fake in-links to a target node (black), whereas manipulating the target's real out-links requires compromising its account and is far harder,  motivating greater trust in out-links than in-links.}
 \label{fig:link-spam}
 \vspace{-0.1in}
\end{wrapfigure}

As evident from the literature~\citep{jin2021adversarial, dai2022comprehensive, mujkanovic2022defenses,hou2024adversarial, hou2024robust}, most existing research on the attack and defense of GNNs focuses on undirected graphs. 
From the \textit{attack perspective}, existing attacks 
~\citep{dai2018adversarial,jin2021adversarial,wu2019adversarial,zugner2018adversarial,zugner2019certifiable} flip both directions of an edge (out-link and in-link) 
, which could be unrealistic in 
real-world scenarios. For instance, for a social network in Fig.~\ref{fig:link-spam}, it is relatively easy to create 
many 
orchestrate large-scale link spam (i.e., in-links) targeting specific users~\citep{alkhalil2021phishing}. However, hacking into the accounts of those target users and manipulating their following behaviors (i.e., out-links) are considerably more difficult~\citep{gohel2015cyber}.
From the \textit{defense perspective}, most robust GNNs~\citep{mujkanovic2022defenses,hou2024adversarial} convert the directed graphs to undirected ones through symmetrization, which neglects valuable directional information.
Despite the existence of directed GNNs, their adversarial robustness is largely unexplored~\cite{tong2020directed,tong2020digraph,zhang2021magnet}. 

In contrast to undirected graphs, many graphs in real-world applications such as citation networks~\citep{radicchi2011citation}, social networks~\citep{robins2009closure}, and web networks~\citep{kleinberg1999web} are naturally directed. The link directions in directed graphs inherently imply trustworthiness: 
\textit{out-links are usually more trustworthy than in-links for a target node}~\citep{page1998pagerank, kamvar2003eigentrust, gyongyi2004combating}.
This is because out-links are usually formed by active behaviors such as citing a paper in citation networks, following a user on social media, linking to a webpage on the Web, or making payment to an account in transaction networks. 
Therefore, it is practically more challenging to attack out-links than in-links of target nodes. 
Regardless of the trustworthiness offered by directed graphs, the fact that most existing attacks and defenses are limited to undirected graphs leaves 
the robustness and trustworthiness of GNNs in directed graphs underexplored.

To address the aforementioned research gap,  we propose to explore adversarial robustness in directed graphs from both attack and defense perspectives for the first time. From the \textit{attack perspective}, we introduce a 
weaker but 
more realistic directed graph attack setting (Section~\ref{sec:rdga}) that differentiates out-link and in-link attacks while imposing certain restrictions on out-link attacks to reflect the practical challenges of manipulating out-links.
From the \textit{defense perspective}, we 
propose a simple yet effective message-passing layer, Biased Bidirectional Random Walk (BBRW),  to protect GNNs against adversarial attacks. 
Our contributions can be summarized as follows:
\vspace{-0.05in}
\begin{itemize}[leftmargin=*]

\item We reveal the potential trustworthiness inherently implied by the link directions in directed graphs, and we propose to explore adversarial robustness 
in directed graphs 
for the first time.

\item We analyze the limitations of existing research on the attacks and defenses in undirected graphs. 
To overcome these limitations,
we introduce Restricted Directed Graph Attack (RDGA) as a more realistic adversarial graph attack 
that differentiates the in-link and out-link attacks on target nodes while restricting the adversary's capability to execute out-link attacks on target nodes.


\item We present an intriguing preliminary study that discovers catastrophic failures of random walk message passing, followed by deeper analyses and understanding. 
This motivates the design of 
a simple yet highly effective
Biased Bidirectional Random Walk (BBRW)
as a plug-in message-passing layer to substantially enhance the robustness of various GNN backbones.

\item Comprehensive comparisons showcase that 
our solution achieves outstanding clean accuracy and state-of-the-art robustness against both transfer and adaptive attacks. 
We provide detailed ablation studies to further understand the working mechanism of the proposed approach. 
\end{itemize}

\section{
Restricted Directed Graph Attack}
\label{sec:rdga}

In this section, we discuss the limitations of existing adversarial graph attacks 
for undirected graphs and introduce a more realistic adversarial graph attack 
for directed graphs. 

\noindent\textbf{Notations.}
In this paper, we consider a directed graph 
$\mathcal{G}=\left(\mathcal{V},\mathcal{E}\right) $ with $|\mathcal{V}|=n$ nodes and $|\mathcal{E}|=m$ edges. The adjacency matrix of $\mathcal{G}$ is denoted as $\vA \in \{0,1\}^{n\times n}$. The feature matrix of $n$ nodes is denoted as $\vX \in \mathbb{R}^{n\times d} $. The label matrix is denoted as $\vY\in \RR^{n}$.
The degree matrix of $\vA$ is $\vD=\text{diag}\left(d_1,d_2,...,d_n\right)$, where $d_i=\sum_j\vA_{ij}$ is the out-degree of node $i$. 
$f_\theta(\vA,\vX)$ denotes the GNN encoder that extracts features from $\vA$ and $\vX$ with parameters $\theta$.

\subsection{Limitations of Existing Adversarial Graph Attack}
\label{sec:existing-attack}
Existing adversarial graph attacks mostly conduct undirected graph attacks that
flip both directions (in-link and out-link) of an adversarial edge once being selected~\citep{xu2019topology,chen2018fast,zugner2018adversarial,zügner2019adversarial}.
However, this common practice has some critical limitations. 
First, it is often impractical to attack both directions of an edge in graphs. For instance, flipping the out-links of users in social media platforms or financial systems usually requires hacking into their accounts to change their following or transaction behaviors, which can be easily detected by security countermeasures such as Intrusion Detection Systems~\citep{bace2001intrusion,hou2024hlogformer}.
Second, the undirected graph attack setting does not distinguish the different roles of in-links and out-links, which fundamentally undermines the resilience of networks. For instance, a large-scale link spam attack targeting a user does not imply the targeted user fully trusts these in-links. But the link spam attack can destroy the feature of target nodes if being made undirected.
Due to these limitations, existing graph attacks are not practical in many real-world applications, and existing defenses can not effectively leverage useful information from directed graphs.


\subsection{Restricted Directed Graph Attack}
\label{sec:directed-attack}

To overcome the limitations of existing attack and defense 
on GNNs, we propose Restricted Directed Graph Attack (RDGA), a more realistic graph attack setting that differentiates between in-link and out-link attacks on target nodes while restricting the adversary’s capability to execute out-link attacks on target nodes, which aligns with the practical challenges of manipulating out-links.

\noindent\textbf{Restricted Directed Graph Attack (RDGA) Setting.}
We denote the directed adversarial attack on the directed graph $\vA\in\{0,1\}^{n\times n}$ as an asymmetric perturbation matrix $\vP \in \{0,1\}^{n\times n}$. The adjacency matrix being attacked is given by: 
$\tA =\vA+(\vone\vone^\top-2\vA)\odot\vP$,
 where $\vone=[1,1,\dots,1]^\top\in\RR^n$ and $\odot$ denotes element-wise product. $\vP_{ij}=1$ means flipping the edge $(i,j)$ (i.e., $\tA_{ij}=0$ if $\vA_{ij}=1$ or $\tA_{ij}=1$ if $\vA_{ij}=0$) while $\vP_{ij}=0$ means keeping the edge $(i,j)$ unchanged (i.e., $\tA_{ij}=\vA_{ij}$). The asymmetric nature of this perturbation matrix indicates the adversarial edges have directions so that one direction will not necessarily imply the attack from the opposite direction.

Given the practical difficulty of attacking the out-links 
, we impose restrictions on the adversary's capacity for executing out-link attacks. 
The Restricted Directed Graph Attack (RDGA) is given by
\begin{align}
\label{eq:restricted_adjacency_perturbation}
\tA&=\vA+(\vone\vone^\top-2\vA)\odot(\vP\odot\vM),
\end{align}
where $\tilde \vP=\vP\odot\vM$ denotes the restricted perturbation.  
When restricting the out-link of nodes $\cT$ (e.g., the target nodes), the mask matrix is defined as
$\vM_{ij}=0 ~\forall i\in\cT, j\in  \cV$ and $\vM_{ij}=1$ otherwise. 
Therefore, we can formalize RDGA as the following optimization problem: 
\begin{equation}
\label{eq:rdga_optimization}
\begin{aligned}
&\argmin_\vP \; \ell_{\text{attack}}\left(f_{\theta^*}(\tA,\vX),\vY \right)\\
\text{s.t.} \; & \theta^* = \argmin_{\theta}\ell_{\text{train}}\left(f_{\theta}(\vA,\vX),\vY \right),\tA=\vA+(\vone\vone^\top-2\vA)\odot(\vP\odot\vM),\|\vP\odot\vM\|_0\leq \Delta.\\
\end{aligned}
\end{equation}
where $\ell_{\text{attack}}$ and $\ell_{\text{train}}$ are the attacking loss and training loss. $\Delta$ is the perturbation budget.

\begin{algorithm}[h!]
\caption{PGD attack under RDGA setting}

\begin{algorithmic}[1] 
\State \textbf{Input:}
initial perturbation $\vP^{(0)}$, budget $\Delta$, learning rate $\eta$, iterations $T$, number of random trials $K$.
\State \textbf{Output:} optimal perturbation $\vP^*$

\For{$t=1,2,...,T$} 
\State Restricted gradient descent: $\vP^{(t)}=\vP^{(t-1)}-\eta\nabla_\vP \ell_{\text{attack}}\odot \vM$
\EndFor

\For{$k=1,2,...,K$} 
\State Draw binary matrix $\vS^{(k)}$ following
$$\vS^{(k)}_{ij}=\left\{
\begin{aligned}
1 \quad & \text{with probability} \; \vP^{(T)}_{ij} \\
0 \quad& \text{with probability} \; 1-\vP^{(T)}_{ij}
\end{aligned}
\right.$$
\EndFor
\State Choose a perturbation $\vP^*$ from $\{\vS^{(k)}\}$ 
yields the smallest attack loss $\ell_{\text{attack}}$ under $\|\vP^*\|_0\leq \Delta$
\end{algorithmic}
\label{alg:pgd-attack}
\end{algorithm}

\noindent\textbf{Attacking Algorithm.}
The proposed RDGA setting provides a more realistic attack budget allocation that differentiates out-link and in-link attacks. To optimize Eq.~\eqref{eq:rdga_optimization} under RDGA setting, we 
first 
perform  restricted gradient descent with the masking matrix $\vM$:
\begin{equation}
\label{eq:restricted_gd}
\vP^{(t)}=\vP^{(t-1)}-\eta\nabla_\vP \ell_{\text{attack}}\odot \vM.
\end{equation}
Then, we sample multiple candidate perturbations based on  $\vP^{(T)}$ and select the strongest  $\vP^*$ among them.
The details of the attacking process are summarized in Algorithm~\ref{alg:pgd-attack}.

Actually, our RDGA setting is compatible with any existing graph attack algorithms including PGD~\cite{xu2019topology}, FGA~\citep{mujkanovic2022defenses}, Nettack~\cite{nettack} and Metattack~\citep{Zgner2019AdversarialAO} by adjusting the attack budgets for out-links and in-links when selecting the edges.
We empirically evaluate the attacking performance on various attacks under RDGA setting in Appendix~\ref{sec:appendix-diff-attacks}. As verified by our experiments, PGD attack exhibits the strongest performance among them, thus we majorly adopt PGD attack  in our main experiment. 
In Section~\ref{sec:ablation}, we also study a more general RDGA that allows some portion of the attack budgets on targets' out-links where the masking matrix is partially masked.

\section{Methodology: Robust Directed GNNs}
\label{sec:method}

While the directed attack proposed in Section~\ref{sec:rdga} is weaker (but more realistic) than existing undirected attacks due to the additional constraints, undirected GNNs can not benefit from such constraints
since they lose directional information after symmetrization. 
In spite of this, it still offers unprecedented opportunities to design robust GNNs 
in directed graphs.


In this section, we provide a preliminary study on GNNs with simple directed random walk message passing that ignores in-links. The catastrophic failures we discover from such a study provide intriguing and valuable insights into designing better solutions.
This motivates the design of simple and effective GNNs with biased bidirectional random walk message passing. We also provide a theoretical case study to 
understand 
the working mechanism of the proposed algorithm.


\subsection{Catastrophic Failures of Directed Random Walk}
\label{sec:rw-mp}

\begin{table}[h!]
    \centering
    \caption{
    Accuracy (\%) on Cora-ML under RDGA. With RW, our GCN-RW and APPNP-RW exhibit great robustness under transfer attack (\textbf{bold}), but suffer from \textit{catastrophic failures} due to indirect attacks under stronger adaptive attacks (\textcolor{red}{red}).}
    \vspace{-0.1in}
    \renewcommand\arraystretch{1.3}
    \small
    \label{tab:directed-propagation}
    \begin{tabular}{cccccccc}
    \toprule
     \multirow{2}{*}{ Model $\backslash$ Budget }&0\%&\multicolumn{2}{c}{25\%}&\multicolumn{2}{c}{50\%}&\multicolumn{2}{c}{100\%}\\ 
    &Clean&Transfer&Adaptive&Transfer&Adaptive&Transfer&Adaptive\\ 
    \hline
        MLP &  73.5±7.4 & 73.5±7.4& 73.5±7.4 & 73.5±7.4 & 73.5±7.4& 73.5±7.4 & 73.5±7.4  \\ 
        \cline{1-8}    
        GCN  & 89.5±6.1 & 66.0±9.7 & 66.0±9.7 & 40.5±8.5 & 40.5±8.5 & 12.0±6.4 & 12.0±6.4  
        \\
        GCN-RW (Ours) &86.5±6.3&\textbf{86.5±6.3}&\textbf{\textcolor{red}{52.0±8.1}}&\textbf{86.5±6.3}&\textbf{\textcolor{red}{28.0±4.6}}&\textbf{86.5±6.3}&\textbf{\textcolor{red}{10.5±5.7}}
        \\
        
        \cline{1-8}
        APPNP &  90.5±4.7& 81.5±9.5 & 80.5±10.4 & 66.5±8.7 &68.0±12.1 & 44.0±9.2 & 46.0±7.3 
        \\ 
        APPNP-RW (Ours) &	85.5±6.5&\textbf{85.5±6.5}&\textbf{\textcolor{red}{30.0±7.7}}&	\textbf{85.5±6.5}&\textbf{\textcolor{red}{15.0±3.9}}	&\textbf{85.0±6.3}&\textbf{\textcolor{red}{11.5±3.2}}
        \\
    
    \bottomrule
    \end{tabular}
\end{table}


Due to the adversary's capacity constraint on out-link attacks, out-links are more reliable than in-links, which aligns better with real-world examples as demonstrated in Section~\ref{sec:intro} and Section~\ref{sec:rdga}. This motivates us to first study a simple directed random walk message passing (RW) that only aggregates node features from out-links: $\vX^{l+1} = \vD^{-1} \vA \vX^l$. 
We use two popular GNNs including GCN~\citep{kipf2016semi} and APPNP~\citep{gasteiger2018predict} as the backbone models and substitute their symmetric aggregation matrix
$\vD^{-\frac{1}{2}} \vA_{\tsym} \vD^{-\frac{1}{2}}$ 
as $\vD^{-1}\vA$, denoted as GCN-RW and APPNP-RW.
We evaluate the clean and robust node classification accuracy of these variants on the Cora-ML dataset under RDGA, following the experimental setting detailed in Section~\ref{sec:exp}.
It is worth emphasizing that while we transfer attacks from the surrogate model GCN as usual, we additionally test the robust performance of adaptive attacks which directly attack the victim model to avoid a potential false sense of robustness. 
The results in Table~\ref{tab:directed-propagation} provide the following insightful observations: 
\begin{itemize}[leftmargin=*]
\item In terms of clean accuracy, we have GCN $>$ GCN-RW $>$  MLP and APPNP $>$ APPNP-RW $>$  MLP. This indicates that both out-links and in-links in the clean directed graph provide useful graph topology information. 
Undirected GNNs (GCN and APPNP) achieve the best clean performance since both in-links and out-links are utilized through symmetrization. 



\item Under transfer attacks, we have GCN-RW $>$ GCN $>$ MLP
and APPNP-RW $>$ APPNP $>$ MLP. 
Transfer attacks barely impact GCN-RW and APPNP-RW since no out-link attack 
is allowed under RDGA setting
and RW is free from the impact of in-link attacks.
However, in-link attacks hurt GCN and APPNP badly due to the symmetrization operation.


\item Although RW performs extremely well under transfer attacks, we surprisingly find that GCN-RW and APPNP-RW suffer from \textit{catastrophic failures} under stronger adaptive attacks, and they significantly underperform simple MLP, which uncovers a severe \textit{false sense of robustness}. 

\item We also test the robustness of three state-of-the-art directed GNNs including 
DGCN~\citep{tong2020directed}, DiGCN~\citep{tong2020digraph},
and MagNet~\citep{zhang2021magnet} under the proposed RDGA setting, but they are still very vulnerable to attacks. We leave these results in Section~\ref{sec:exp} (Table~\ref{tab:acc-cora}) to avoid duplication. 


\end{itemize}

\begin{figure}[h!]
 \centering
 
 \includegraphics[width=0.7\textwidth]{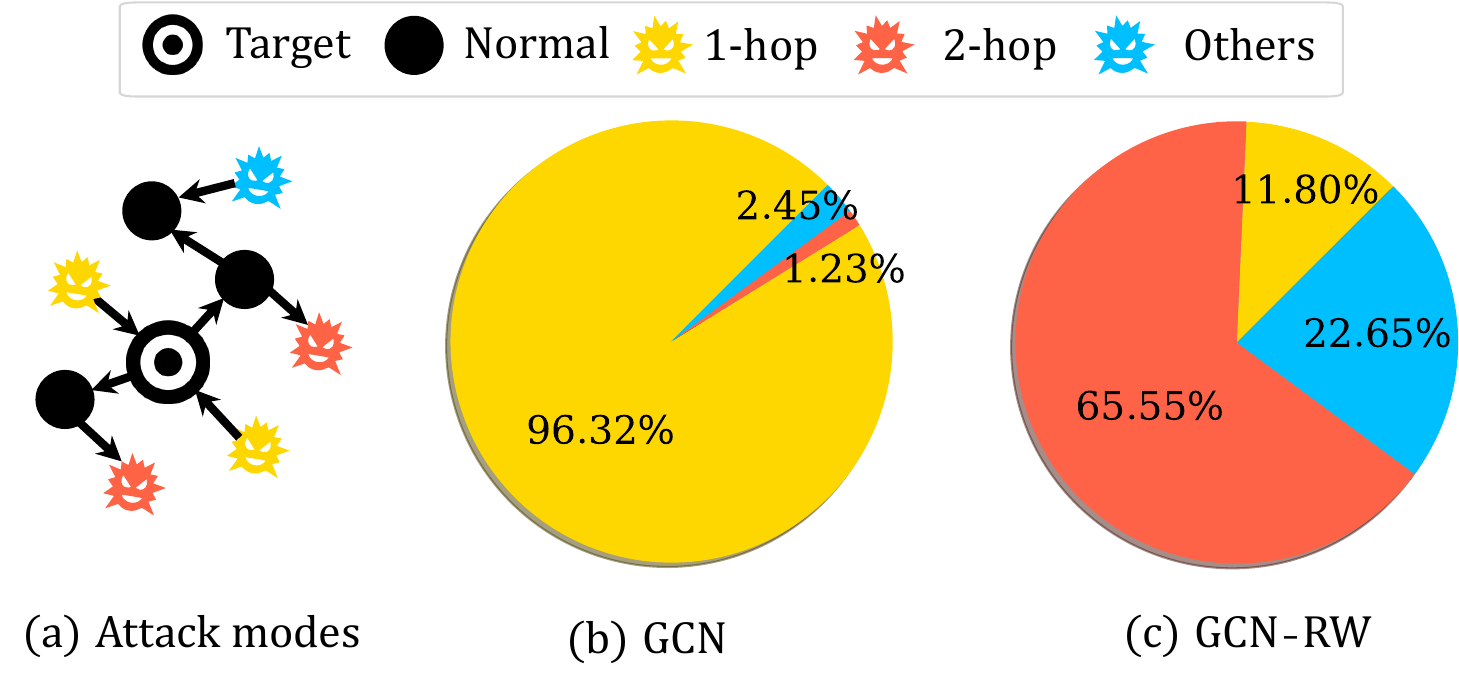}
 \vspace{-0.15in}
 \caption{Adversary behaviors. (a) Attack modes defined relative to the target node. (b) Attacking GCN: the adversary concentrates on direct 1-hop in-link attacks (96.32\%). (c) Attacking GCN-RW: direct in-link attacks are largely ineffective, so the adversary shifts to indirect 2-hop attacks via the target's out-linking neighbors (65.55\%), exposing GCN-RW's false sense of robustness.
 }
 \label{fig:secondary-attack-stat}
 \vspace{-0.1in}
\end{figure}

\noindent\textbf{Catastrophic Failures due to Indirect Attacks.} 
The {catastrophic failures} of GCN-RW and APPNP-RW under adaptive attacks indicate their {false sense of robustness}.  
In order to understand this phenomenon and gain deeper insights,
we perform statistical analyses on the adversary behaviors when attacking different victim models such as GCN and GCN-RW using attack budget $50\%$ (Fig.~\ref{fig:secondary-attack-stat}). 
Note that similar observations can be made under other attack budgets as shown in Appendix~\ref{sec:appendix-supp-figure}. 
In particular, we separate adversarial links into different groups according to whether they directly connect target nodes or targets' neighbors.
The yellow portion represents 
attacks by 1-hop neighbors on the target (direct in-link attacks); the red portion
represents attacks by 2-hop neighbors on the target 
(indirect out-link
attacks); and the blue portion represents other attacks. 
The distributions 
of adversarial links shown in Fig.~\ref{fig:secondary-attack-stat} indicate:
\begin{itemize}[leftmargin=*]
\item 
When attacking GCN (Fig.~\ref{fig:secondary-attack-stat} (b)), the adversary majorly attacks the 1-hop in-links of target nodes using 96.32\% perturbation budget, which badly hurts its performance since GCN relies on both in-links and out-links. However, the attack transferred from these two victim models barely impact GCN-RW that only trusts out-links.

\item 
When attacking GCN-RW (Fig.~\ref{fig:secondary-attack-stat} (c)), the adversary can not manipulate the 1-hop out-links of target nodes under the restricted setting (RDGA). 
It does not focus on attacking the 1-hop in-links of target nodes either since these 1-hop in-links can not influence GCN-RW.
Instead, the adversary tactfully identifies the targets' neighbors and conducts 2-hop out-link attacks through these neighbors using 65.55\% budget. 
In other words, it focuses on attacking out-linking neighbors of target nodes so these neighbors can destroy the predictions of target nodes.
\end{itemize}

\subsection{Biased Bidirectional Random Walk } 
\label{sec:birw-mp}
The study on the directed random walk message passing in Section~\ref{sec:rw-mp} indicates that it is non-trivial to robustify GNNs using directed graphs, but it provides insightful motivations to develop a better approach. In this section, we propose a simple and effective approach with theoretical justification.
The systematic study on Section~\ref{sec:rw-mp} offers two valuable lessons: \textbf{(1)} Both in-links and out-links provide useful graph topology information; 
\textbf{(2)} While out-links are more reliable than in-links, full trust in out-links can cause catastrophic failures and a false sense of robustness under adaptive attacks due to the existence of indirect attacks. 
These lessons motivate us to develop a message-passing framework that 
fully utilizes the out-links and in-links information and
differentiates their roles. Importantly, it also needs to avoid a false sense of robustness under adaptive attacks.

To this end, we propose a simple yet highly effective Biased Bidirectional Random Walk (BBRW)
Message Passing framework 
that balances the trust on out-links and in-links:
$$\tilde \vA_\beta = \vD^{-1}_{\beta} \vA_\beta,\text{where } \vD_{\beta} = \vA_\beta \vone,
\vA_\beta=\beta \vA + (1-\beta) \vA^\top.$$
$\vA_\beta$ is the weighted sum of $\vA$ and $\vA^\top$ that combines the out-links (directed random walk) and in-links (inversely directed random walk), i.e., $\{\vA_\beta\}_{ij} = \beta \vA_{ij} + (1-\beta) \vA_{ji}$. $\vD_{\beta}$ is the out-degree matrix of $\vA_\beta$. $\tilde \vA_\beta$ denotes the random walk normalized propagation matrix that aggregates node features from both out-linking and in-linking neighbors.

\noindent\textbf{Bias weight.} The bias weight $\beta\in[0,1]$ controls the relative trustworthiness of out-links compared with in-links. 
When $\beta=1$, it reduces to RW that fully trusts out-links. But RW suffers from catastrophic failures under adaptive attacks as shown in Section~\ref{sec:rw-mp}. Therefore, $\beta$ is typically recommended to be selected in the range $(0.5,1)$ to reflect the assumption that out-links are more reliable than in-links but out-links are not fully trustworthy due to the existence of indirect in-link attacks on the neighbors. However, if we believe in-links are more reliable than out-links, BBRW provides the flexibility to take the advantage of such prior information by choosing $\beta \in (0, 0.5)$. 


\noindent\textbf{Advantages.}
The proposed BBRW enjoys the advantages of simplicity, trustworthiness, explainability, universality, and efficiency.
First, BBRW is simple due to its clear motivation and easy implementation
with only one hyperparameter. 
Second, the hyperparameter $\beta$ provides the flexibility to adjust the trust between out-links and in-links, which helps avoid catastrophic failures and the false sense of robustness caused by the unconditional trust in out-links.
The working mechanism 
of this hyperparameter are clearly justified by a theoretical analysis in Section~\ref{sec:theoretical-analysis}.
Moreover, it can be readily used as a plug-in layer to improve the robustness of various GNNs, 
and it shares the same computational and memory complexities as the backbone GNNs. BBRW is also orthogonal and compatible with existing defense strategies developed for undirected GNNs.

\subsection{Theoretical Analysis of BBRW}
\label{sec:theoretical-analysis}


\begin{figure}[h!]
\begin{center}
\subfigure[]{
\includegraphics[width=0.3\columnwidth]{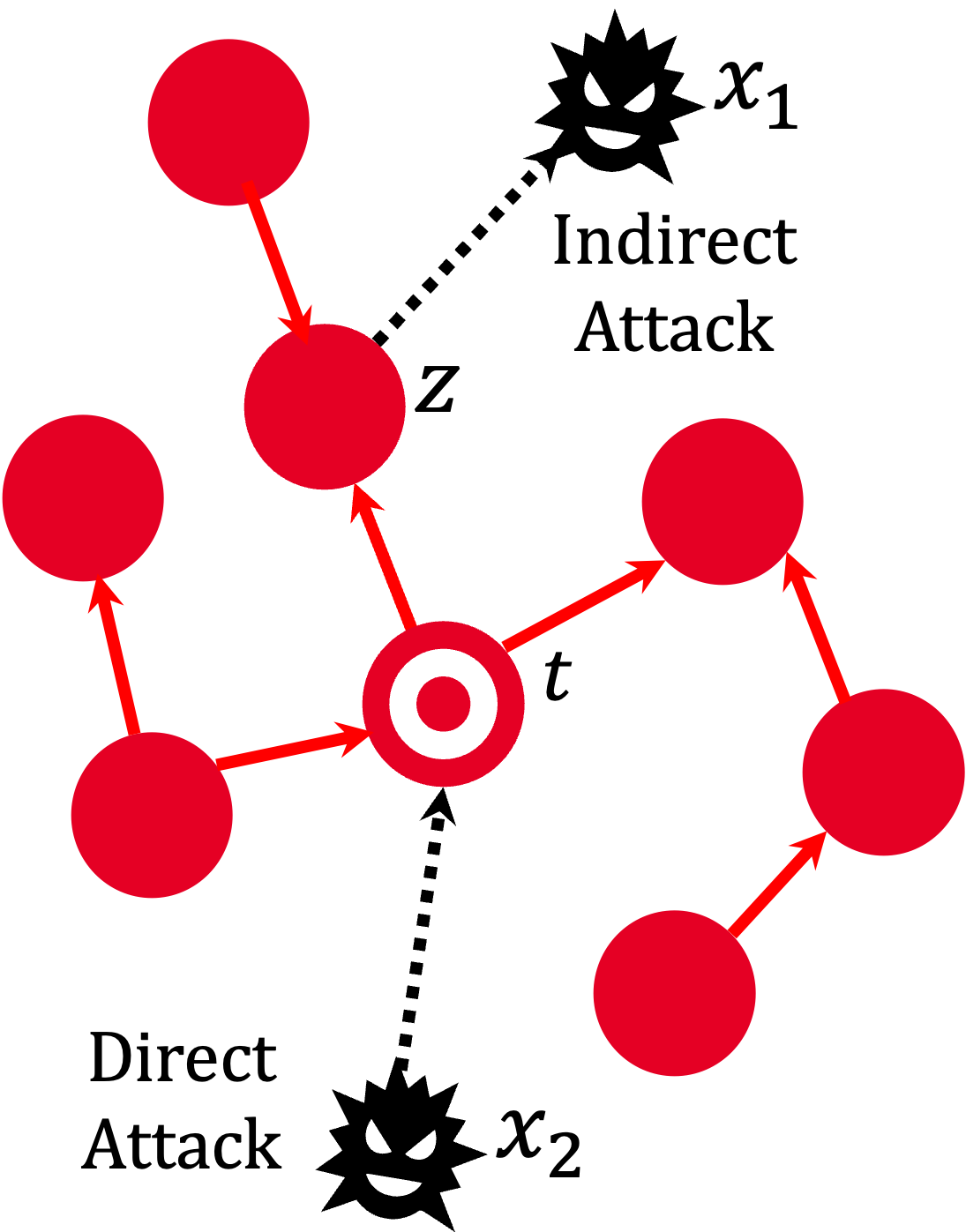}}
\subfigure[]{\includegraphics[width=0.5\columnwidth]{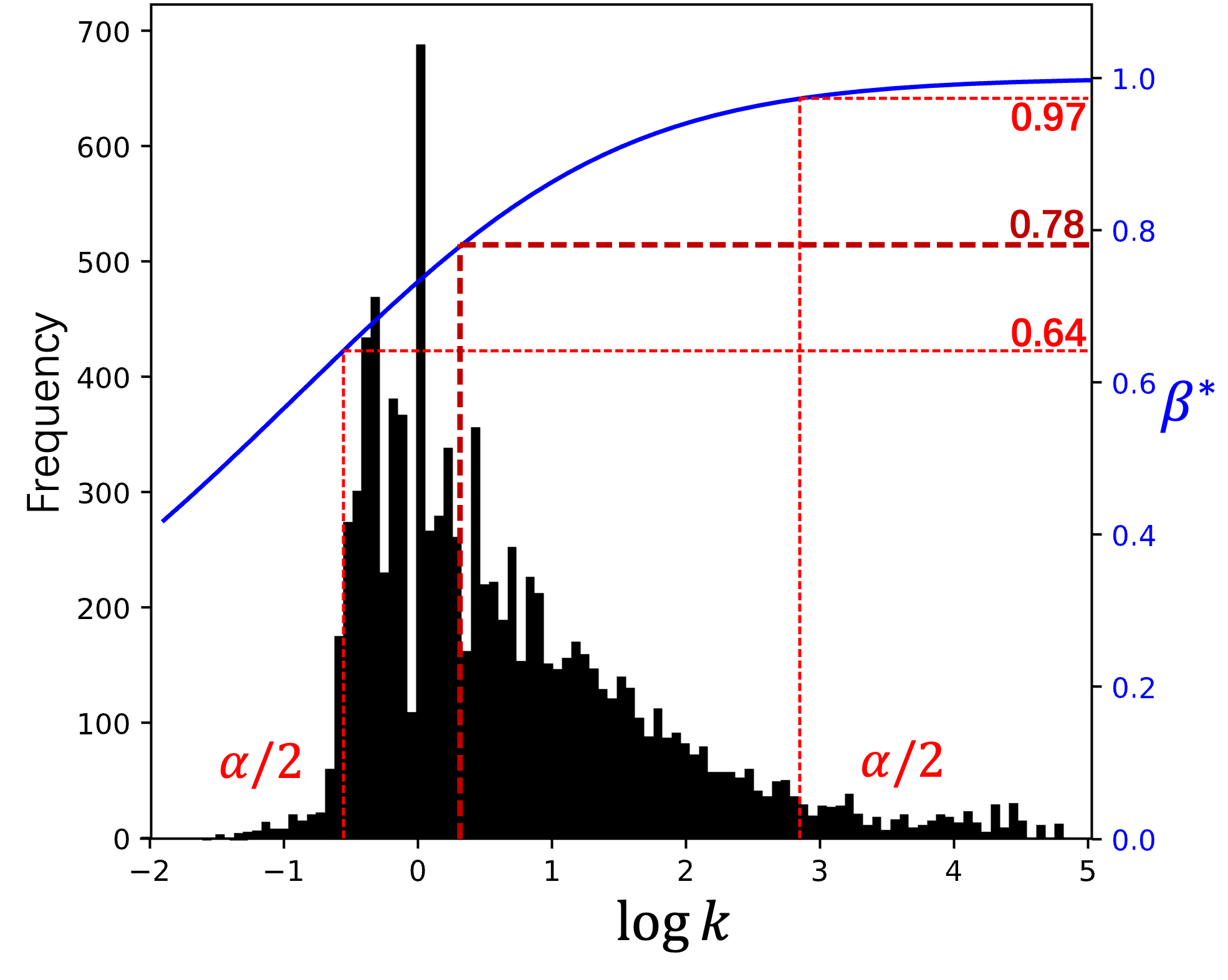}}
\end{center}
\vspace{-0.15in}
\caption{Theoretical analysis of BBRW. (a) Direct attack ($x_2\ to t$) vs. indirect attack ($x_1\to z \to t$) considered in Theorem A.1, which derives the optimal bias weight $\beta^*$ that minimizes the worse-case attack influence. (b) Distribution of $\log k$ over 10,000 sampled node triplets from Cora-ML (black histogram) and the corresponding theoretical $\beta^*(k)$ (blue curve). The median $\beta^*=0.79$ with 80\% confidence interval $(0.68, 0.92)$, matching the empirically optimal $\beta$ in Section 4.3.}
\label{fig:analysis-RDGA}

\end{figure}

We provide a theoretical analysis of BBRW to understand its working mechanism.
Let $h_x^{(0)}$ and $h_x^{(k)}$ be the input feature and the $k$-th layer hidden feature of node $x$ in GNNs. The
influence score  
$I(x,y)=\left\|\frac{\partial h_x^{(k)}}{\partial h_y^{(0)}}\right\|_1$
can measure the impact of node $y$ on node $x$ in the message passing~\citep{xu2018representation}.
The attack mechanisms of out-link indirect attack and in-link direct attack are shown in Fig.~\ref{fig:analysis-RDGA} (a).
In our hyperparameter settings, we employ 2 layer neural networks for BBRW, resulting in  $2$-step random walk message passing: $\tilde \vA_\beta^2$. 
Therefore, the increment of influence score after conducting 
the attack on node $t$ is equivalent to $\tilde \vA_\beta^2(t,x_1)$  for \emph{indirect attack} or $\tilde \vA_\beta^2(t,x_2)$ for \emph{direct attack}.
From the perspective of defense, we need to robustify the GNNs by choosing an appropriate $\beta$ to mitigate the potential impact of the stronger attack among them:
\begin{align}
\label{eq:min-max-attack}
    \beta^*&=\argmin_{\beta\in [0,1]}\max\{\tilde \vA_\beta^2(t,x_1),\tilde \vA_\beta^2(t,x_2)\}.
\end{align}

{
\renewcommand{\thetheorem}{\ref{thm:theory-beta}}
\begin{theorem}
Define the factor 
$k:=\frac{\vD^{-1}_\beta(t)+\vD^{-1}_\beta(x_2)}{2\vD^{-1}_\beta(z)}$, then the theoretical optimal $\beta^*$ in Eq.~\eqref{eq:min-max-attack} is given by $\beta^*(k)=\sqrt{k^2+2k}-k$.
\end{theorem}
\addtocounter{theorem}{-1}
}

For a target node $t$, Theorem~\ref{thm:theory-beta} provides the theoretical optimal $\beta^*$ that minimizes the attack influence from a 1-hop direct attack or a 2-hop indirect attack through a neighbor node $z$.
The detailed proof is presented in Appendix~\ref{sec:proof-theorem}.
Since different nodes may have a different $\beta^*$ due to their distinct $k$, we perform a statistical analysis 
on the distribution of $\beta^*$. From Cora ML, we randomly select node $t$, $x_2$, and its neighbor $z$ such that we calculate and collect
multiple samples $\{k^{(i)}\}_{i=1}^{10000}$.
Fig.~\ref{fig:analysis-RDGA} (b) shows the histogram of $\{\log k^{(i)}\}_{i=1}^{10000}$ and  $\beta^*$  in terms of $\log k$.
From the figure, we can find that the optimal $\beta^*$ has a median of $0.79$ 
and the $1-\alpha$ ($\alpha=0.2$) confidence interval is $(0.68,0.92)$. This optimal value range
aligns well with the optimally tuned $\beta^*$ in the ablation study in Section~\ref{sec:ablation} (Fig.~\ref{fig:ablation-beta-cora} and Fig.~\ref{fig:ablation-beta-cite}). This further substantiates the validity of our approach.


\section{Experiment}
\label{sec:exp}

In this section, we provide comprehensive experiments to verify the advantages of the proposed BBRW. Comprehensive ablation studies are presented to illustrate its working mechanism. 

\subsection{Experimental Setting}
\label{sec:exp-setting}

\textbf{Datasets.} We mainly use the two most widely used datasets in the literature, namely Cora ML~\citep{mccallum2000automating} and Citeseer~\citep{giles1998citeseer}. Additionally, we also include three larger citation datasets: Ogbn-Arxiv~\citep{hu2021open}, PubMed~\citep{sen2008collective} and WikiCS~\citep{mernyei2022wikics}.  Beyond citation graphs, we evaluate our approaches in other four datasets including AmazonPhoto~\citep{shchur2018pitfalls}, Flickr~\citep{zeng2019graphsaint}, AmazonComputers~\citep{shchur2018pitfalls}, and Reddit~\citep{hamilton2017inductive}.
We use the following data splits: 10\% training, 10\% validation, and
80\% testing.
We repeat the experiments for 10 random data splits and report the mean and variance of the node classification accuracy.

\noindent\textbf{Baselines.}  We compare our models with several undirected GNNs:
GCN~\citep{kipf2016semi}, APPNP~\citep{gasteiger2018predict},
Jaccard-GCN~\citep{wu2019adversarial}, RGCN~\citep{zhu2019robust}, GRAND~\citep{feng2020graph}, GNNGuard~\citep{zhang2020gnnguard}, SoftMedian~\citep{geisler2021robustness}, HANG~\citep{zhao2024adversarial}, NoisyGNN~\citep{ennadir2024simple},  GARNET~\citep{deng2022garnet}, and RUNG~\citep{hou2024robust},
most of which are designed as robust GNNs. Additionally, we also select three state-of-the-art directed GNNs including 
DGCN~\citep{tong2020directed}, DiGCN~\citep{tong2020digraph}
and MagNet~\citep{zhang2021magnet}. 



\noindent\textbf{Hyperparameter settings.} 
For all methods, hyperparameters are tuned from the following search space: 1) learning rate: \{0.05, 0.01, 0.005\};
2) weight decay: \{5e-4, 5e-5, 5e-6\}; 3) dropout rate:
\{0.0, 0.5, 0.8\}. For APPNP, we use the teleport probability $\alpha=0.1$ and propagation step $K=10$ as \citep{gasteiger2018predict}. For BBRW, we tune $\beta$ in $[0,1]$ with the interval 0.1.
For a fair comparison, the proposed BBRW-based methods share the same architectures and hyperparameters with the backbone models.
For all models, we use 2 layer neural networks with 64
hidden units. 

\noindent\textbf{Adversary attacks \& evaluations.} 
We mainly conduct evasion target attacks using PGD topology attack~\citep{xu2019topology} under the proposed RDGA setting. The details of the attacking algorithm are presented in Algorithm~\ref{alg:pgd-attack}. 
We randomly select 20 target nodes per split for robustness evaluation and run the experiments for multiple link budgets $\Delta \in \{0\%, 5\%, 10\%, 15\%, 20\%, 25\%, 50\%, 100\%\}$ of the target node's total degree. Additionally, we also evaluate the robustness under the global attack, node injection attack~\citep{tao2021single} and influencer attack~\citep{zugner2020adversarial}. Furthermore, we also include the certified robustness evaluation to provide a provable and theoretical robustness for our method.
In the experimental results,
\emph{Transfer} and \emph{Adaptive} refer to transfer and adaptive attacks, respectively. For transfer attacks, we choose a 2-layer GCN as the surrogate model following existing works~\citep{mujkanovic2022defenses,zugner2018adversarial}. For adaptive attacks, the victim models are the same as the surrogate models, 
avoiding a false sense of robustness in transfer attacks. 
``$\backslash$" means we do not find a trivial solution for adaptive attack since it is non-trivial to compute the gradient of the adjacency matrix for those victim models. 

\subsection{Robust Performance}

\begin{table}[!ht]
    \centering
    \caption{ Classification accuracy (\%) under different perturbation rates of graph attack. 
    The best results are in \textbf{bold}, and the second-best results are \underline{underlined}. (Cora-ML)}
    \vspace{-0.1in}
    \renewcommand\arraystretch{1.3}
    \label{tab:acc-cora}
    \resizebox{0.99\linewidth}{!}{%
    \begin{tabular}{cccccccc}
    \toprule
     \multirow{2}{*}{Method}&0\%&\multicolumn{2}{c}{25\%}&\multicolumn{2}{c}{50\%}&\multicolumn{2}{c}{100\%}\\
    &Clean &Transfer&Adaptive&Transfer&Adaptive&Transfer&Adaptive\\ \hline
        MLP &  73.5±7.4 & 73.5±7.4 & 73.5±7.4 & 73.5±7.4 & 73.5±7.4 & 73.5±7.4 & 73.5±7.4 \\ 
        \cline{1-8}
        DGCN&89.5±7.6&76.5±13.0&$\backslash$&54.5±7.9&$\backslash$&38.0±14.2&$\backslash$\\ 
    DiGCN& 85.0±7.4 &50.0±6.7&$\backslash$&40.5±9.1& $\backslash$ &29.0±6.2 &$\backslash$\\ 
    MagNet&88.5±3.2 &70.5±10.6&$\backslash$&59.5±10.6&$\backslash$&54.0±7.0&$\backslash$\\
    \cline{1-8}
Jaccard-GCN&90.5±6.5&69.5±7.9&65.5±7.9&44.0±6.2&34.0±7.0&21.0±7.0&8.0±4.6\\ 
        RGCN&88.0±6.0&72.5±8.4&66.0±7.7&44.0±8.9&36.0±5.4&17.5±8.7&7.0±4.6\\ 
GRAND&85.5±6.1&74.0±7.0&65.0±7.4&64.0±9.2&51.0±8.6&45.0±7.1&24.0±7.7\\ 
        GNNGuard &90.0±5.0&87.5±6.4&75.0±8.7&82.5±7.2&61.0±7.3&75.0±8.4&28.0±3.3\\
 GARNET &
 90.5$\pm$6.5 &
 79.0$\pm$7.5 &
 76.0$\pm$7.5 &
 72.0$\pm$8.0 &
 70.0$\pm$8.0 &
 50.0$\pm$9.5 &
 48.0$\pm$9.5 \\

 HANG &
 91.0$\pm$7.0 &
 80.0$\pm$7.5 &
 77.5$\pm$7.5 &
 73.0$\pm$6.5 &
 71.0$\pm$6.5 &
 51.0$\pm$3.5 &
 49.0$\pm$3.5 \\

 NoisyGNN &
 90.0$\pm$5.5 &
 76.5$\pm$6.0 &
 74.0$\pm$6.0 &
 70.5$\pm$6.0 &
 68.0$\pm$6.0 &
 49.0$\pm$4.5 &
 47.0$\pm$4.5 \\

 RUNG &
 92.0$\pm$5.5 &
 88.0$\pm$7.0 &
 85.0$\pm$7.0 &
 78.0$\pm$8.5 &
 76.0$\pm$7.5 &
 52.0$\pm$11.0 &
 50.0$\pm$9.5 \\

        \cline{1-8}
        GCN &  89.5±6.1 & 66.0±9.7 & 66.0±9.7 & 40.5±8.5 & 40.5±8.5 & 12.0±6.4 & 12.0±6.4  \\ 
        BBRW-GCN (Ours)&90.0±5.5 &\underline{89.5±6.1} 
 &\underline{89.0±6.2} &\underline{86.0±5.4} &\underline{85.0±6.3}&\underline{85.0±7.1} &\underline{75.0±10.2}     \\
        \cline{1-8}
        APPNP &  90.5±4.7 & 81.5±9.5 & 80.5±10.4 & 66.5±8.7 &66.0±7.9 & 44.0±9.2 & 43.5±6.4  \\ 
        BBRW-APPNP (Ours)& 91.0±4.9  &89.0±5.4 &87.5±5.6 & 85.0±7.1 &83.0±6.4 &83.5±6.3 &69.0±9.7 \\ \cline{1-8}
        SoftMedian &\underline{91.5±5.5} & 86.0±7.0 & 83.0±7.1 & 75.0±8.4 & 73.0±7.1 & 48.5±11.4 & 47.5±9.3\\ 
        BBRW-SoftMedian 
 (Ours)& \textbf{92.0±4.6} &\textbf{91.5±5.0} &\textbf{92.0±4.6} 
        & \textbf{89.5±6.9} &\textbf{88.0±5.1} & \textbf{87.0±8.4}  &\textbf{84.5±8.8} \\ \bottomrule
    \end{tabular}
    } 
\end{table}

\begin{table}[!ht]
    \centering
    \caption{
    Classification accuracy (\%) under different perturbation rates of graph attack. The best results are in \textbf{bold}, and the second-best results are \underline{underlined}. (Citeseer)}

    \renewcommand\arraystretch{1.0}
    \label{tab:acc-cite}
    \resizebox{0.99\linewidth}{!}{%
    \begin{tabular}{cccccccc}
    \toprule
     \multirow{2}{*}{Method}&0\%&\multicolumn{2}{c}{25\%}&\multicolumn{2}{c}{50\%}&\multicolumn{2}{c}{100\%}\\ 
    &Clean &Transfer&Adaptive&Transfer&Adaptive&Transfer&Adaptive\\\midrule

        MLP & 49.0±9.4 & 49.0±9.4 & 49.0±9.4 & 49.0±9.4 & \textbf{49.0±9.4} & \textbf{49.0±9.4} & \textbf{49.0±9.4}\\ 
         \cline{1-8}
    DGCN&64.0±7.0&54.0±8.3&$\backslash$&34.5±10.6&$\backslash$&27.0±10.1&$\backslash$\\ 
    DiGCN&66.0±8.6&41.5±10.5&$\backslash$&29.5±8.2&$\backslash$&21.5±5.9&$\backslash$\\ 
    MagNet&68.0±6.0&51.5±11.2&$\backslash$&35.0±12.0&$\backslash$&35.0±7.7&$\backslash$\\
    \midrule
    Jaccard-GCN&57.0±7.1&45.5±7.9&38.5±9.5&23.0±7.8&11.5±5.5&20.0±10.2&6.5±5.0\\
    RGCN&61.5±7.1&34.5±9.1&34.0±10.2&9.5±4.2&7.0±5.6&6.5±4.5&4.5±3.5\\
    GRAND&67.5±6.0&56.5±6.3&56.0±8.9&43.0±5.1&42.5±9.0&37.5±8.1&27.5±6.8\\
        GNNGuard&60.5±7.2&50.0±8.7&43.5±9.0&33.0±8.7&18.0±8.4&31.5±8.7&8.5±3.9\\
 GARNET &
 62.0$\pm$7.0 &
 52.5$\pm$8.0 &
 50.0$\pm$8.0 &
 36.0$\pm$9.0 &
 34.0$\pm$8.5 &
 28.0$\pm$9.0 &
 27.0$\pm$8.5 \\

 HANG &
 63.0$\pm$7.5 &
 53.5$\pm$8.0 &
 51.5$\pm$8.0 &
 37.0$\pm$8.5 &
 35.5$\pm$8.0 &
 29.0$\pm$8.0 &
 28.0$\pm$7.5 \\

 NoisyGNN &
 61.0$\pm$6.5 &
 51.0$\pm$7.5 &
 49.0$\pm$7.5 &
 35.0$\pm$8.0 &
 33.0$\pm$7.5 &
 27.0$\pm$8.0 &
 26.0$\pm$7.5 \\

 RUNG &
 64.5$\pm$6.5 &
 57.5$\pm$8.0 &
 55.0$\pm$8.0 &
 38.5$\pm$9.0 &
 37.0$\pm$8.5 &
 31.0$\pm$9.0 &
 30.0$\pm$8.5 \\

    \midrule
        GCN&59.0±5.4&36.5±9.5&36.5±9.5&10.5±5.7&10.5±5.7&4.5±4.2&4.5±4.2\\
        BBRW-GCN (Ours)&61.5±7.4 &50.0±7.7 &43.0±10.3 &31.5±6.3 &27.0±14.4 &26.0±8.0 &20.5±9.6 \\
        \midrule
        APPNP&\textbf{72.0±6.0}&53.5±9.5&51.0±6.2&16.0±10.7&13.5±98&9.0±4.4&8.5±9.0\\
        BBRW-APPNP (Ours)&\underline{69.0±4.4}&\textbf{66.0±8.3} &\textbf{59.0±9.7}&\textbf{55.0±8.1} &26.5±8.4 &\underline{43.5±6.3} &14.5±6.1 \\
        \midrule
        SoftMedian &61.5±5.9&56.0±8.3&56.0±8.3&34.5±10.8&35.0±10.7&26.5±9.8&26.0±9.0\\
        BBRW-SoftMedian (Ours) &59.5±7.2 &\underline{58.5±7.8} &\underline{58.5±7.8} &\underline{53.0±7.5} &\underline{48.0±7.0} &\textbf{49.0±7.7} &\underline{48.0±8.1} \\
        \bottomrule
    \end{tabular}
    } 
\end{table}

To demonstrate the
benefits
of the proposed BBRW message-passing framework, we develop multiple variants of it by plugging BBRW into classic GNN backbones: GCN~\citep{kipf2016semi}, APPNP~\citep{gasteiger2018predict}, and SoftMedian~\citep{geisler2021robustness}.
The clean and robust performance with attack budgets \{25\%, 50\%, 100\%\} are compared with plenty of representative GNN baselines on Cora-ML and Citeseer datasets as summarized in Table~\ref{tab:acc-cora} and Table~\ref{tab:acc-cite}, respectively. We leave the results for low budgets \{5\%, 10\%, 15\% 20\%\} in Table~\ref{tab:acc-cora-low-budget} and Table~\ref{tab:acc-citeseer-low-budget}  in Appendix~\ref{sec:results_low_budget}.
From these results, we can observe the following: 

\begin{itemize}[leftmargin=*]

\item In most cases, all baseline GNNs underperform the graph-agnostic MLP under adaptive attacks, which indicates their incapability to robustly leverage graph topology information. However, most BBRW variants outperform MLP. Taking Cora-ML as an instance, the best BBRW variant (BBRW-SoftMedian) significantly outperforms MLP by $\{18\%, 16\%, 13.5\%\}$ (transfer attack) and $\{18.5\%, 14.5\%, 11\%\}$ (adaptive attack) under $\{25\%, 50\%, 100\%\}$ attack budgets. Even under 100\% perturbation, BBRW-SoftMedian still achieves 84.5\% robust accuracy under strong adaptive attacks, which suggests the value of directional information.

\item The proposed BBRW is a highly effective plug-in layer that significantly and consistently enhances the robustness of GNN backbones in both transfer and adaptive attack settings. Taking Cora-ML as an instance, under increasing attack budgets $\{25\%, 50\%, 100\%\}$: (1) BBRW-GCN outperforms GCN by $\{23.5\%,$ $ 45.5\%, 73\%\}$ (transfer attack) and $\{23\%, 44.5\%, 63\%\}$ (adaptive attack); (2) BBRW-APPNP outperforms APPNP by $\{7.5\%, 18.5\%,$ $ 39.5\%\}$ (transfer attack) and $\{7\%, 17\%, 25.5\%\}$ (adaptive attack); (3) BBRW-SoftMedian outperforms SoftMedian by $\{5.5\%, 14.5\%,$ $38.5\%\}$ (transfer attack) and $\{9\%, 15\%, 37\%\}$ (adaptive attack). The improvements are stronger under larger attack budgets. 


\item The proposed BBRW not only significantly outperforms existing directed GNNs such as DGCN, DiGCN, and MagNet in terms of robustness but also exhibits consistently better clean accuracy. BBRW also 
outperforms existing robust GNNs under attack. Compared with undirected GNN backbones such as GCN, APPNP, and SoftMedian, BBRW maintains 
comparable clean accuracy.

\end{itemize}

\subsection{Ablation Study}
\label{sec:ablation}

In this section, we further perform ablation studies on the attacking patterns, hyperparameter setting, and adversary capacity in RDGA to understand the working mechanism of the proposed BBRW. Additionally, we also evaluate the robustness under various scenarios (including global attack, certified robustness, node injection attack, and larger dataset) to further validate its effectiveness. 

\begin{wrapfigure}{r}{0.6\textwidth}
 \centering
 \includegraphics[width=0.43\textwidth]{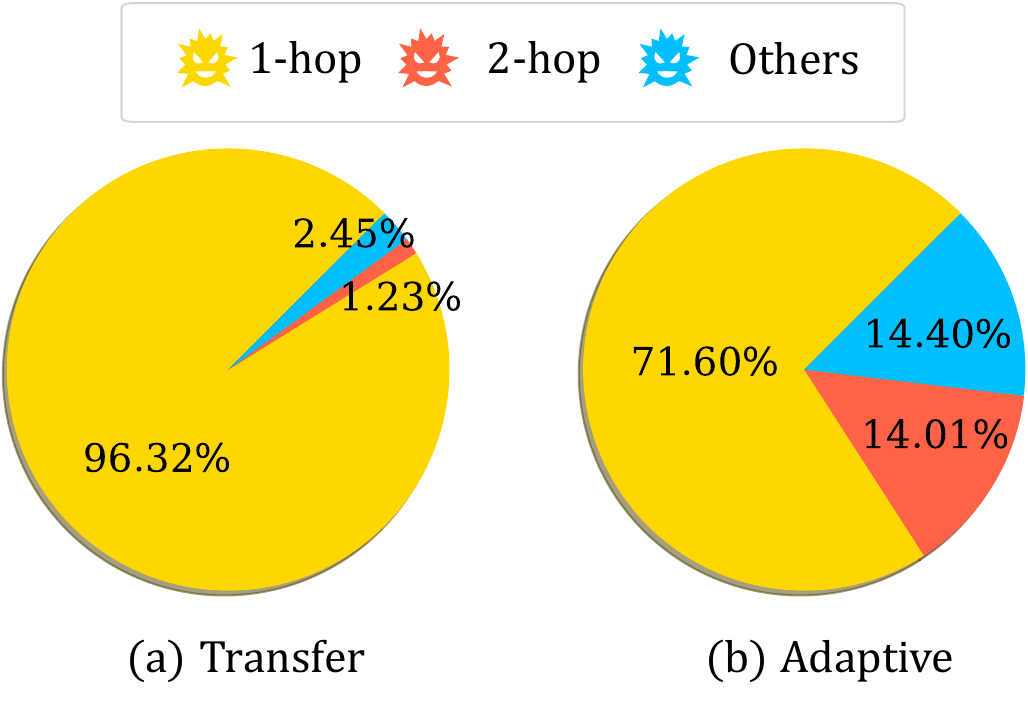}
 
    \vspace{-0.15in}
       \caption{Adversary behaviors against BBRW-SoftMedian ($\beta=0.7$). (a) Transfer: adversary focuses on direct 1-hop in-link attacks (96.32\%), barely affecting BBRW-SoftMedian. (b) Adaptive: adversary shifts budget to indirect 2-hop (14.01\%) and other attacks, yet accuracy stays at 87\%.
       }
    
         \label{fig:attack-patterns-BBRW-SM}
 \vspace{-0.2in}
\end{wrapfigure}

\noindent\textbf{Attacking patterns.} In Table~\ref{tab:acc-cora}, we observe that BBRW- SoftMedian overwhelmingly outperform all baselines in terms of robustness. To investigate the reason, we show the adversarial attack patterns of transfer and adaptive attacks on BBRW-SoftMedian ($\beta=0.7$) in Fig.~\ref{fig:attack-patterns-BBRW-SM}.
In the transfer attack, the adversary spends 96.32\% budget on in-links attacks on 
 the target nodes directly, which causes a minor effect on BBRW-SoftMedian that trusts out-links more. In the adaptive attack, 
the
adversary is aware of the biased trust of BBRW and realizes that in-links attacks are not sufficient. Therefore, besides direct in-link attacks, it allocates 14.01\% and 14.40\% budgets to conduct the out-links indirect attacks on targets' neighbors and other attacks. Even though the adversary optimally adjusts the attack strategy, BBRW-SoftMedian still achieves an incredible 87\% and 84.5\% robust accuracy under 50\% and 100\% total attack budgets. This verifies BBRW's extraordinary capability to defend against adaptive attacks.

\begin{figure}[h!]
 \centering
\includegraphics[width=0.99\textwidth]{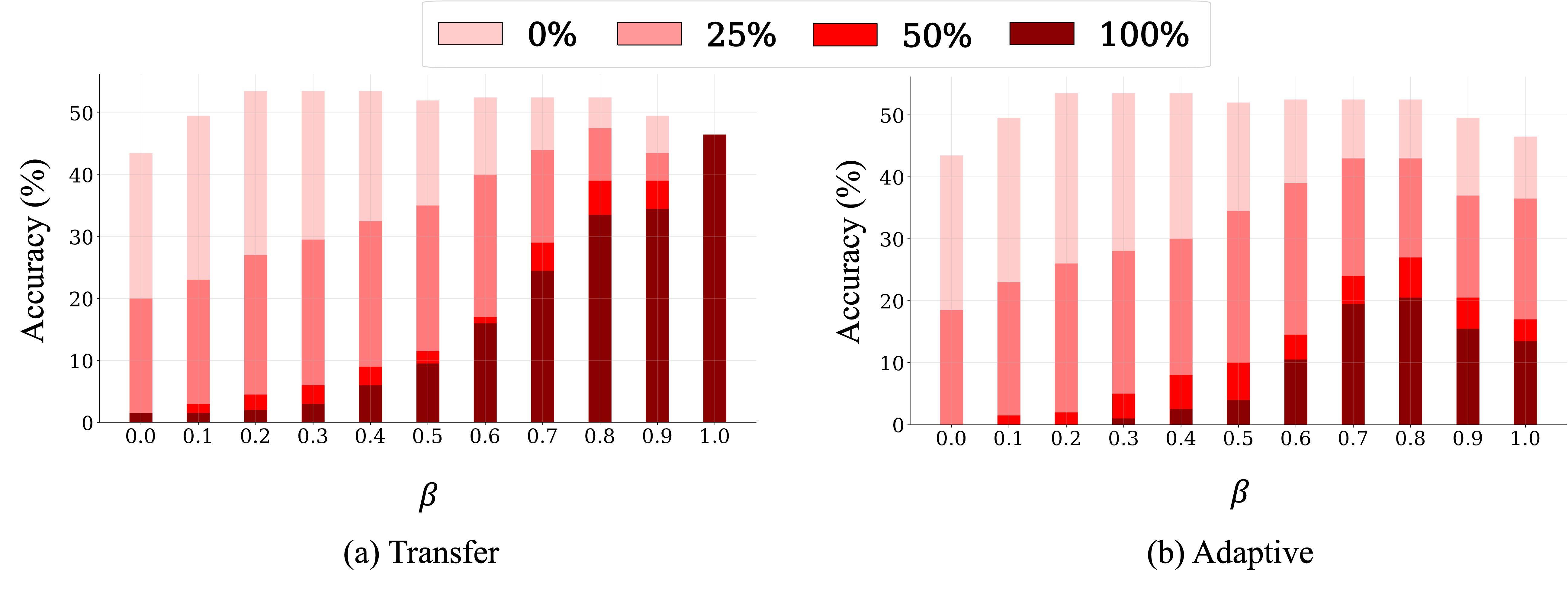}
 \vspace{-0.15in}
\caption{Ablation study of $\beta$ on Citeseer under transfer and adaptive attacks. Colors denote classification accuracy at different attack budgets. Increasing $\beta$ generally improves robustness under transfer attacks. Under adaptive attacks, performance peaks around $\beta=0.7$--$0.8$ and declines as $\beta$ approaches $1$, particularly under larger attack budgets, further demonstrating the need to balance the trust assigned to in-links and out-links.}
   \label{fig:ablation-beta-cite}
\end{figure}

\noindent\textbf{Hyperparameter in BBRW.} BBRW is a simple and efficient approach. The only hyperparameter is the bias weight $\beta$ that provides the flexibility to differentiate and adjust the trust between out-links and in-links. We study the effect of $\beta$ by varying $\beta$ from 0 to 1 with an interval of 0.1 using BBRW-GCN.
The accuracy under different attack budgets on Cora-ML and Citeseer is summarized in Fig.~\ref{fig:ablation-beta-cora} and Fig.~\ref{fig:ablation-beta-cite}, respectively. 
We can make the following observations:

\begin{figure}[h!]
 \centering
 \includegraphics[width=0.99\textwidth]{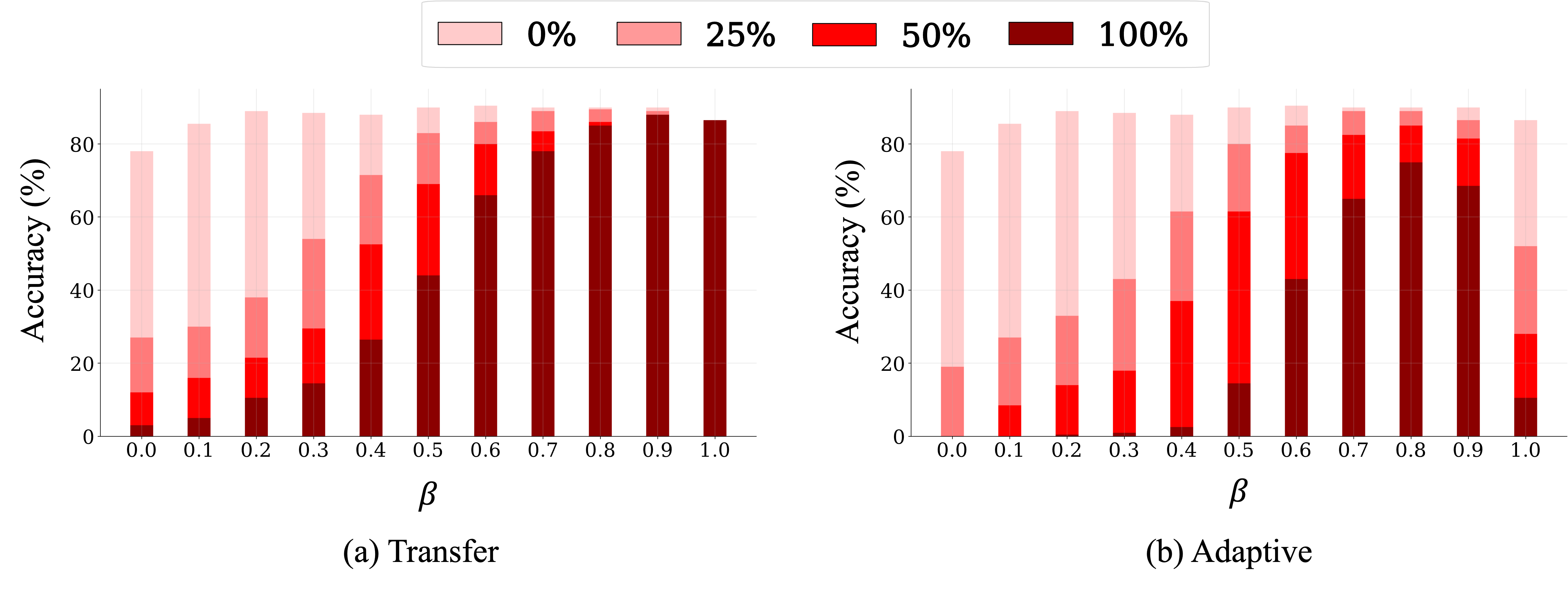}
 \vspace{-0.15in}
\caption{Ablation study of $\beta$ on Cora-ML under transfer and adaptive attacks. Colors denote classification accuracy at different attack budgets. Increasing $\beta$ generally improves robustness under transfer attacks. Under adaptive attacks, accuracy increases with $\beta$, peaks around $\beta=0.8$, and then drops at $\beta=1$, especially under larger attack budgets, highlighting the importance of balancing the trust assigned to in-links and out-links.}
 \label{fig:ablation-beta-cora}
 \vspace{-0.1in}
\end{figure}

\begin{itemize}[leftmargin=*]


\item Under transfer attacks, BBRW-GCN becomes more robust with the growth of $\beta$. It demonstrates that larger $\beta$ reduces the trust and impact of in-link attacks on target nodes.

\item Under adaptive attacks, BBRW-GCN becomes more robust with the growth of $\beta$ but when it transits to the range close to $\beta=1$ (GCN-RW), it suffers from catastrophic failures due to the indirect out-link attacks on targets' neighbors, which is consistent with the discovery in Section~\ref{sec:rw-mp}. 
This also explains the false sense of robustness evaluated under transfer attacks.


\end{itemize}

\noindent\textbf{Adversary capacity in RDGA.}
One of the major reasons BBRW can achieve extraordinary robustness 
is to differentiate the roles and trust of in-links and out-links. In RDGA, we assume that the adversary can not manipulate the out-links of target nodes by fully masking target nodes' out-links (i.e., masking rate=100\%).
However, in reality, it is beneficial to consider more dangerous cases when the adversary may be able to manipulate some proportion of targets' out-links. 
Therefore, we also provide ablation study on the general RDGA setting by varying the masking rates of targets' out-links from 0\% to 100\%. The total attack budget including in-links and out-links is set as 50\% of the degree of the target node. The ablation studies on masking rate under adaptive attack and transfer attack are shown in Fig.~\ref{fig:ablation_all} (a) and Fig.~\ref{fig:masking_rate_transfer} in Appendix~\ref{sec:appendix-supp-figure}.
The results in Fig.~\ref{fig:ablation_all} (a) offer the following observations: 
(1) The robustness of undirected backbone GNNs is not affected by constraints  on the out-link attacks 
, as they can't differentiate the out-links and in-links;
(2) BBRW can significantly enhance the robustness of backbones models 
under varying masking rates. The improvements are stronger when out-links are better protected (higher mask rate).

\begin{figure}[h!]
\begin{center}
\subfigure[Masking rate.]{\includegraphics[width=0.325\columnwidth]{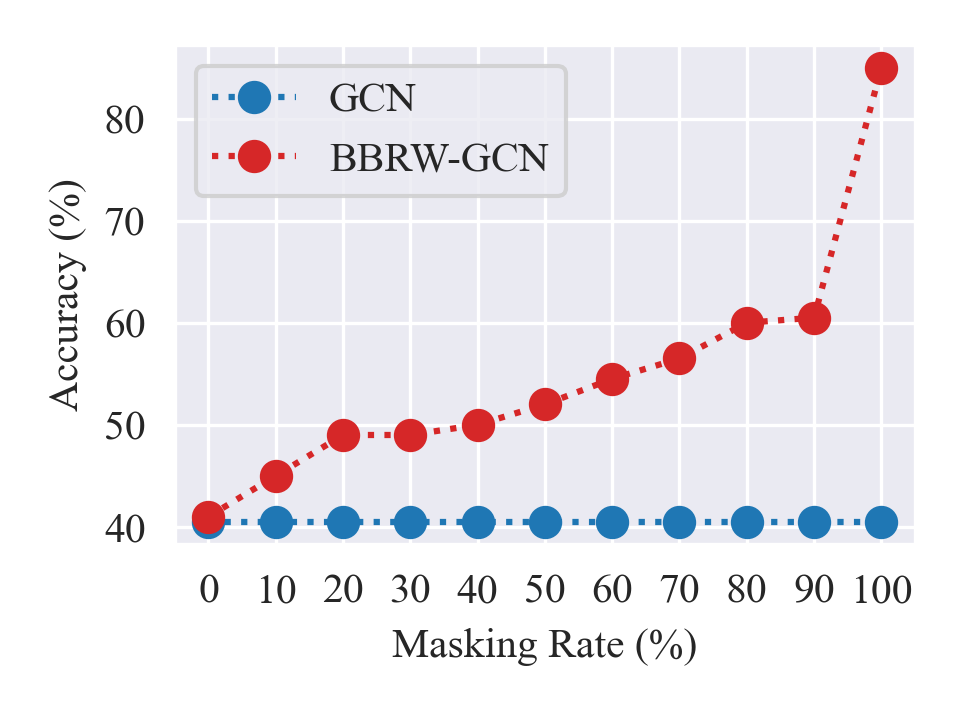}}
\subfigure[Global attack.]{\includegraphics[width=0.325\columnwidth]{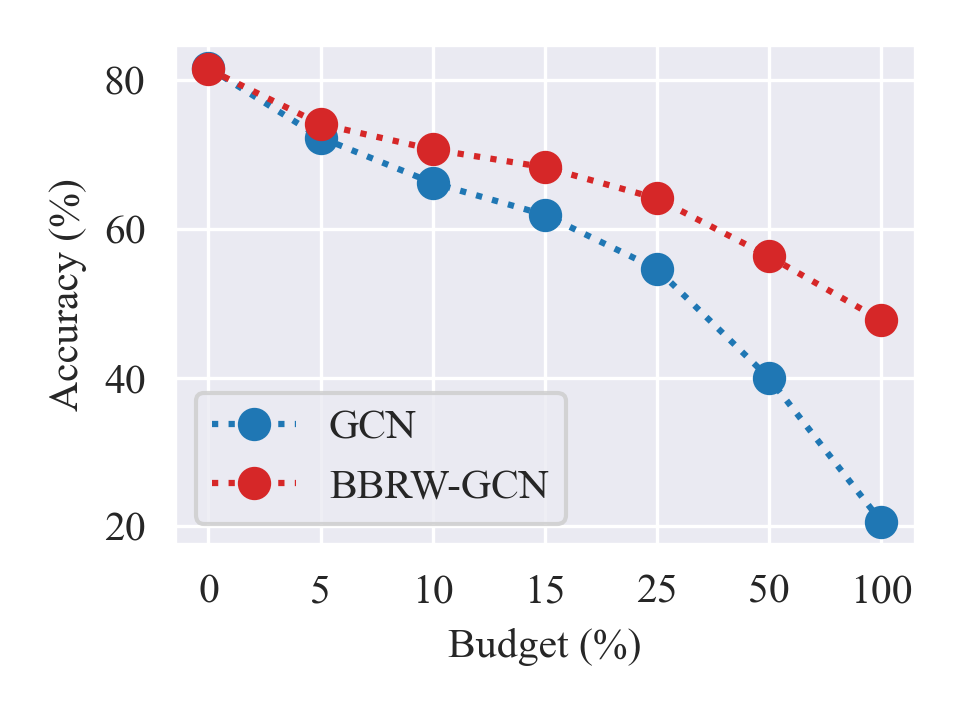}}
\subfigure[Certified robustness.]{\includegraphics[width=0.325\columnwidth]{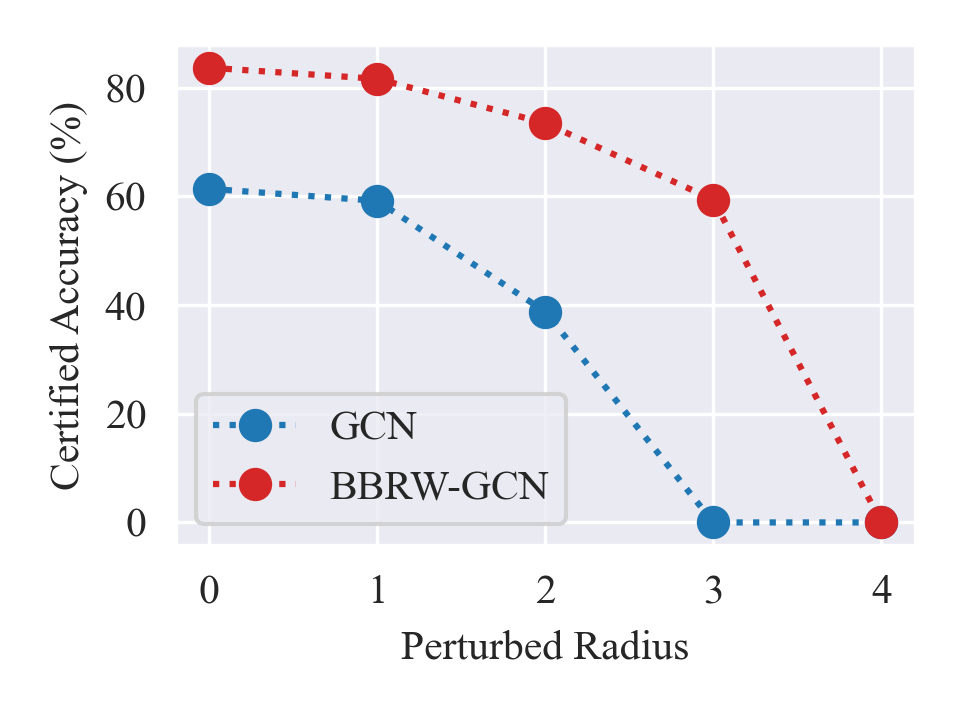}}
\subfigure[Robustness on larger citation datasets.]{\includegraphics[width=0.68\columnwidth]{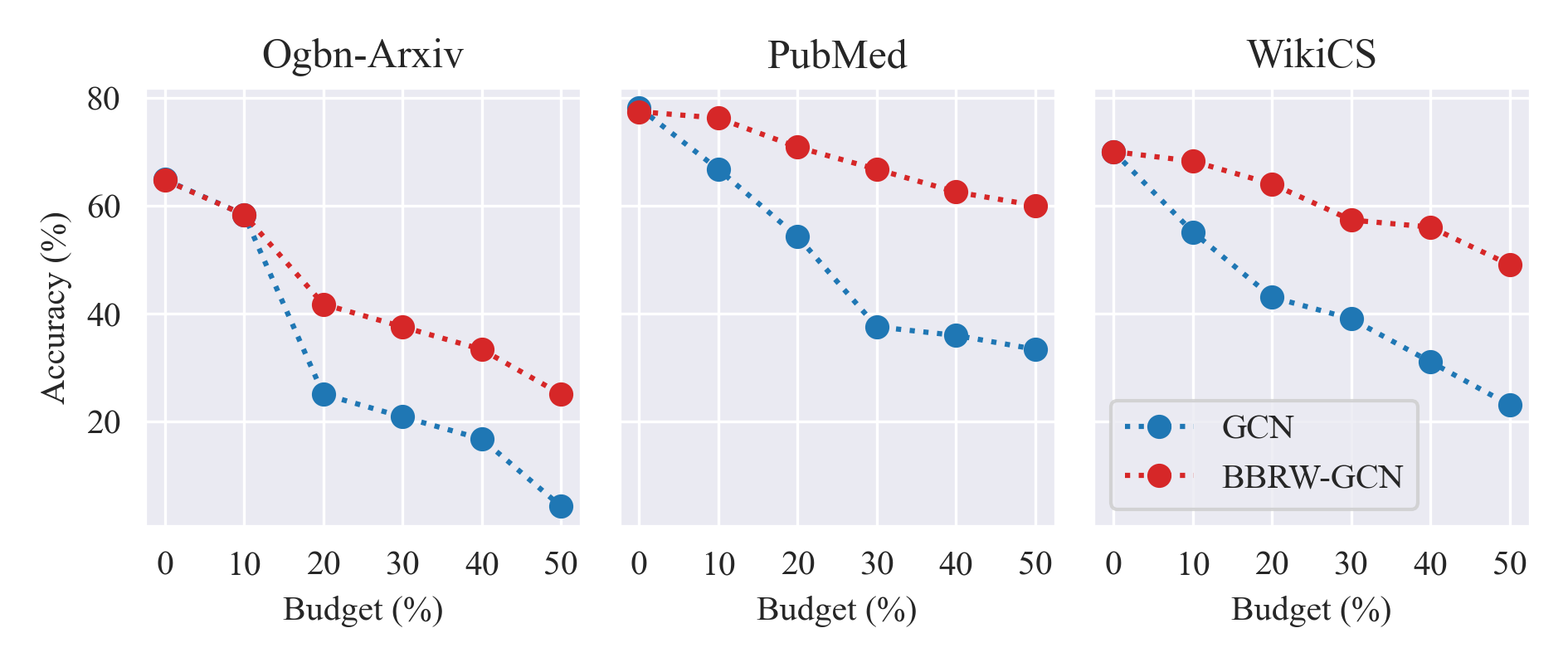}}
\subfigure[Node injection attack. ]{\includegraphics[width=0.30\columnwidth]{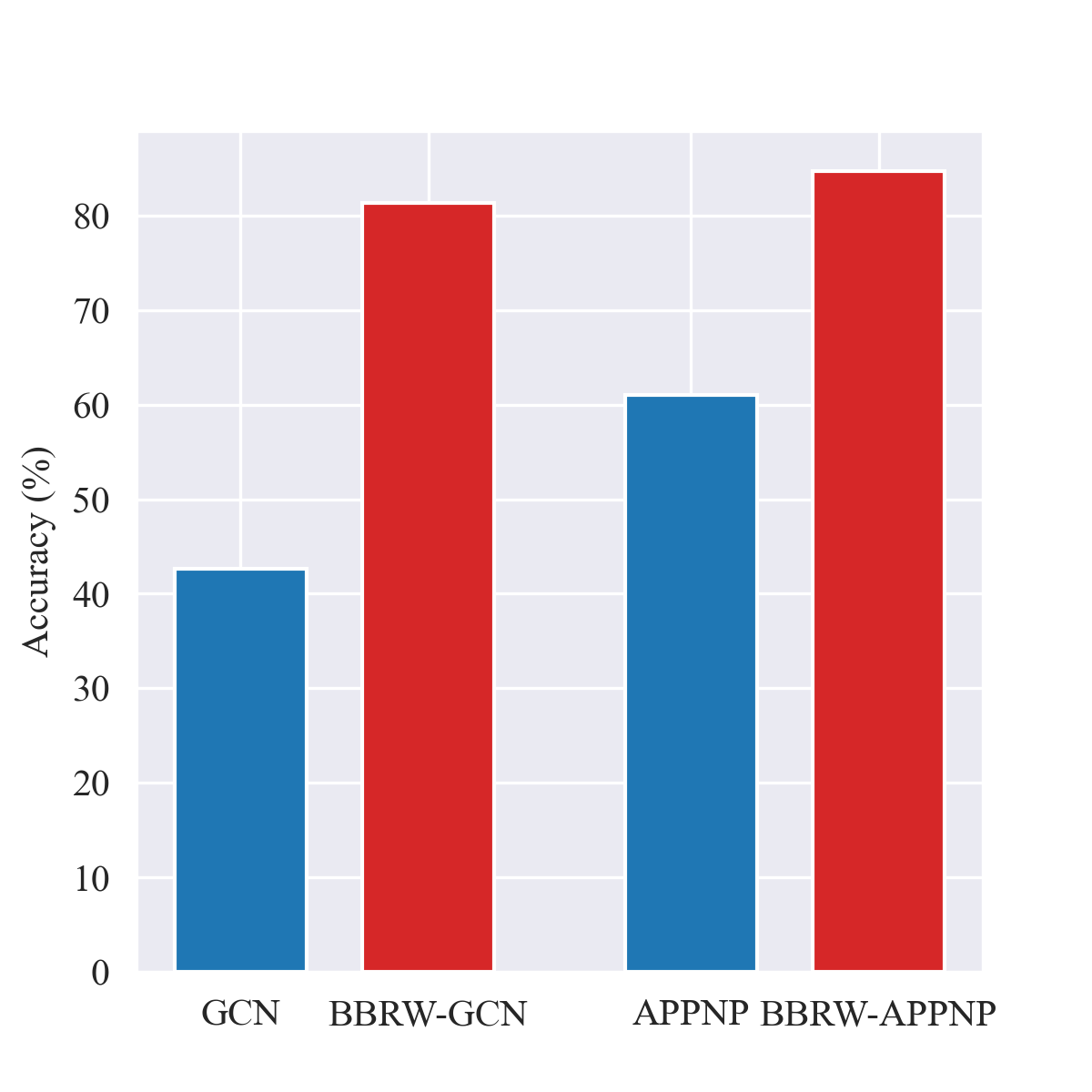}}
\subfigure[Datasets beyond citation graphs.]{\includegraphics[width=0.99\columnwidth]{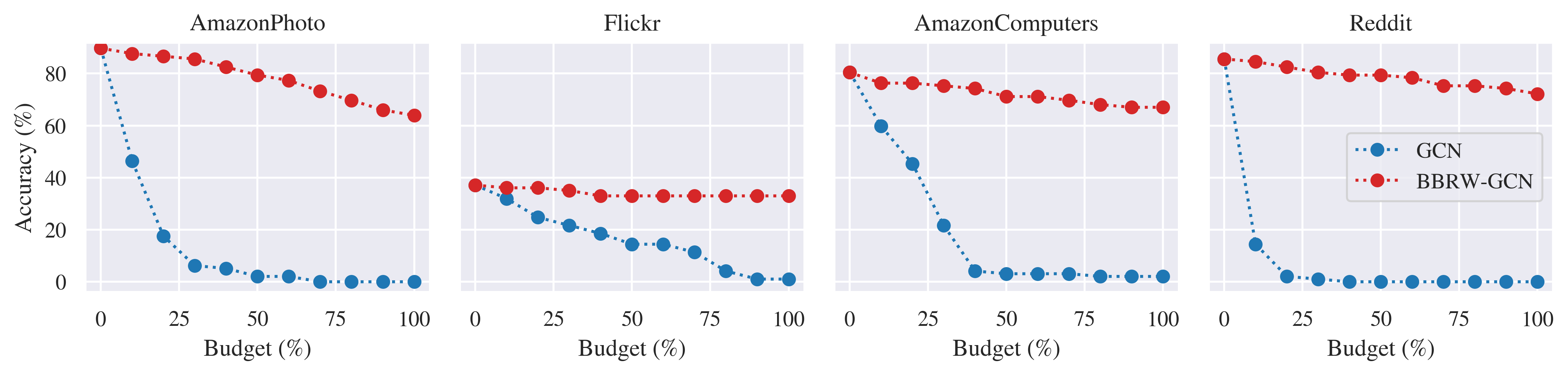}}
\end{center}
\caption{Robustness evaluation across various attack settings and datasets. (a) BBRW provides stronger robustness as more target out-links are protected. (b) BBRW consistently improves the backbone models under global attacks. (c) BBRW achieves stronger certified robustness across different perturbation budgets. (d) BBRW remains effective on larger citation networks with more complex structures. (e) BBRW mitigates the impact of adversarial edges introduced through node injection. (f) BBRW consistently improves robustness on datasets beyond citation networks.}
\label{fig:ablation_all}
\end{figure}

\noindent\textbf{Global attack.} 
Besides local attacks, our method also improves the robustness under global attacks. In global RDGA, we mask the out-links perturbations of the test nodes rather than the specific target node. The results in Fig.~\ref{fig:ablation_all} (b) validate that BBRW-GCN can significantly outperform the GCN backbone across various budgets, especially 
under the larger budgets.

\noindent\textbf{Certified robustness.} The above experiments have demonstrated the significant effectiveness of our BBRW under the empirical adversarial attack. 
Furthermore, we leverage the randomized smoothing on GNNs in~\cite{wang2021certified} evaluate the certified robustness can provide a more reliable evaluation with theoretical guarantee regardless of the attack. 
The following experimental results in Fig.~\ref{fig:ablation_all} (c) show that our BBRW-GCN can significantly outperform the backbone GCN in
certified robustness.

\noindent\textbf{Large-scale citation datasets.} 
Our BBRW is a simple and efficient method which shares the same scalability as the backbone GNN models. To further validate the effectiveness of our BBRW on the large scale dataset, we include another three datasets: Ogbn-Arxiv, Pubmed and WikiCS. The budget is set as 0\%-50\% of the target node's degree. The results in Fig.~\ref{fig:ablation_all} (d) show that our proposed method exhibit much better robustness across multiple datasets.

\noindent\textbf{Node injection attack.} 
In addition to the classical network perturbation, we also consider the node injection attacks, which injects malicious nodes rather than modifying original  edges in the existing graph. We utilize the G-NIA~\cite{tao2021single} to conduct the injection attack on the backbone models and our BBRW methods, the following results in 
Fig.~\ref{fig:ablation_all} (e)
show that
our BBRW can differentiate the direction of injected edges and significantly mitigate the effect of the directed injected edges.

\noindent\textbf{Datasets beyond citation graphs.}
To further examine the generalization ability of our BBRW beyond citation-style benchmarks, we additionally evaluate it on four non-citation graph datasets, including AmazonPhoto, Flickr, AmazonComputers, and Reddit. These datasets cover diverse real-world scenarios such as co-purchase networks, social networks, and large-scale online communities
, and thus pose different structural challenges. 
Following the same experimental protocol, the attack budget is varied from 0\% to 100\% of the target node's degree. As shown in Fig.~\ref{fig:ablation_all} (f), BBRW 
achieves significantly higher accuracy than the backbone GCN under increasing perturbation budgets. In particular, while the performance of GCN degrades rapidly even under small budgets, BBRW exhibits much slower performance decay and maintains stable accuracy across all datasets, demonstrating its strong robustness and effectiveness in a wide range of graph structures.

\begin{wraptable}{r}{0.45\textwidth}
 \vspace{-0.02in}

\caption{Accuracy (\%) of BBRW-GCN under RDGA and indirect attacks. RDGA consistently results in lower accuracy as the attack budget increases, demonstrating its greater attack strength.}
    \centering
 \vspace{-0.1in}
    \renewcommand\arraystretch{1.2}
    \label{tab:indriect}
    \resizebox{0.99\linewidth}{!}{%
    \begin{tabular}{cccccccc}

    \toprule
    &0\% &25\%&50\%&100\%\\ \hline
        RDGA&  61.5±7.4 &43.0±10.3 &27.0±14.4 &20.5±9.6 \\  
        Indirect&61.5±7.4 &60.0±8.4 &46.0±10.7&35.0±7.1\\
        \bottomrule
    \end{tabular}
    } 
 \vspace{-0.1in}
\end{wraptable}

\noindent\textbf{RDGA and indirect attack.}
 Our proposed attack RDGA can be seen as a bridge of direct attack and indirect attack. As shown in Fig.~\ref{fig:analysis-RDGA}, the attacker tries to make a decision to maximize the attack strength between in-link direct attack and indirect attack. 
    Therefore, RDGA is stronger than the indirect (adaptive) attack. To verify this conclusion, we also provide the comparison of the strengths of RDGA and Indirect adaptive attack  with BBRW-GCN in Citeseer as in Table~\ref{tab:indriect}. 
    The comparison indicates that the proposed RDGA attack is indeed stronger than the Indirect attack setting by~\cite{zugner2018adversarial} .

\vspace{-0.1in}
\section{Related Work}
\label{sec:relate}



Existing research on the attacks and defenses of GNNs focuses on undirected GNNs that convert the graphs into undirected graphs~\citep{chen2018fast,zügner2019adversarial,zugner2018adversarial,xu2019topology,zhu2019robust,zhang2020gnnguard,feng2020graph,jin2020graph,entezari2020all,geisler2021robustness, zhao2024adversarial, ennadir2024simple,  deng2022garnet, hou2024robust, jin2022empowering, li2022reliable, li2024boosting, chen2025adedgedrop, zhuang2025refine, gosch2023adversarial, zhang2023minimum, huang2023robust, gosch2023revisiting, zhang2025can}. 
Therefore, these works can not fully leverage the rich directed link information in directed graphs. 
There exist multiple directed GNNs designed for directed graphs but the robustness is largely unexplored.
The work~\citep{ma2019spectral} proposes a spectral-based GCN for directed graphs by constructing a directed Laplacian matrix using the random walk matrix and its stationary distribution.
DGCN~\citep{tong2020directed} extends spectral-based graph convolution to directed graphs by utilizing first-order and second-order proximity.
MotifNet~\citep{monti2018motifnet} 
uses convolution-like anisotropic graph filters based on local sub-graph structures called motifs.
DiGCN~\citep{tong2020digraph} proposed a directed Laplacian matrix based on the PageRank matrix.
MagNet~\citep{zhang2021magnet} utilizes a complex Hermitian matrix called the magnetic Laplacian to encode undirected geometric structures in the magnitudes and directional information in the phases. 
GNNGuard~\citep{zhang2020gnnguard} introduces a robust propagation through reweighting and can be potentially extended to direct graphs, but it does not leverage the directional information to enhance robustness.
The BBRW proposed in this work is a general framework that can equip various GNNs with the superior capability to handle directed graphs more effectively.

\vspace{-0.1in}
\section{Conclusion}
\label{sec:con}


This work conducts a novel exploration of GNN robustness in the context of directed graphs from both attack and defense perspectives. In contrast to existing studies that primarily focus on undirected settings, we show that directed graphs contain additional structural cues, where the directionality of links implicitly encodes valuable trustworthiness information. Our study indicates that such direction-aware trustworthiness offers unprecedented opportunities to safeguard graph learning models against adversarial attacks by enabling more selective and reliable message propagation. This key observation is validated through a simple yet highly effective solution—biased bidirectional random walk—which explicitly incorporates link direction when aggregating neighborhood information, thereby mitigating the influence of malicious perturbations.
We conduct comprehensive robustness evaluations under a wide spectrum of attack scenarios to rigorously assess the effectiveness of the proposed approach. Experimental results demonstrate that our method can be seamlessly employed as a plug-in message-passing layer to substantially enhance the robustness of various GNN backbones, including models that are already considered robust. Notably, this improvement is achieved without altering the core architecture or training objective of the underlying models. Overall, the findings of this work reveal a promising and previously underexplored research direction for robust graph learning. 


\vspace{-0.1in}
\begin{acks}
Zhichao Hou and Xiaorui Liu are supported by the National Science Foundation (NSF) under grant numbers IIS-2443182 and CMMI-2609644.
\end{acks}

\bibliographystyle{ACM-Reference-Format}
\bibliography{sample-base}

\appendix

\newpage
\section{Proof of Theorem}
\label{sec:proof-theorem}
\begin{theorem}
\label{thm:theory-beta}
Define 
$k :=\frac{\vD^{-1}_\beta(t)+\vD^{-1}_\beta(x_2)}{2\vD^{-1}_\beta(z)}$, then the theoretical optimal $\beta^*$ in Eq.~\eqref{eq:min-max-attack} is given by $\beta^*(k)=\sqrt{k^2+2k}-k$.
\end{theorem}

\begin{proof}
The increment of influence score $\Delta I$ after conducting the indirect attack on node $t$ is equivalent to:
$$\tilde \vA_\beta^2(t,x_1)=\tilde \vA_\beta(t,z)\cdot \tilde\vA_\beta(z,x_1)=\frac{\beta}{\vD_\beta(t)}\cdot \frac{\beta}{\vD_\beta(z)}
,$$ and the increment of influence score after applying  a direct attack on node $t$ will be proportional to:
\begin{align*}
\tilde \vA_\beta^2(t,x_2)&=\tilde \vA_\beta(t,t)\cdot \tilde\vA_\beta(t,x_2)+\tilde \vA_\beta(t,x_2)\cdot \tilde\vA_\beta(x_2,x_2)\\
&=\frac{1}{\vD_\beta(t)}\cdot \frac{1-\beta}{\vD_\beta(t)}+\frac{1-\beta}{\vD_\beta(t)}\cdot \frac{1}{\vD_\beta(x_2)}.
\end{align*}

According to Eq.~\eqref{eq:min-max-attack} and the definition of $k$, we have: 
\begin{align*}
    \beta^*&=\arg\min_{\beta\in [0,1]}\max\{\tilde \vA_\beta^2(t,x_1),\tilde \vA_\beta^2(t,x_2)\}\\
    &=\arg\min_{\beta\in [0,1]}\max\{    \frac{\beta}{\vD_\beta(t)}\cdot \frac{\beta}{\vD_\beta(z)},\frac{1}{\vD_\beta(t)}\cdot \frac{1-\beta}{\vD_\beta(t)}+\frac{1-\beta}{\vD_\beta(t)}\cdot \frac{1}{\vD_\beta(x_2)}\}
    \\
    &=\arg\min_{\beta\in [0,1]}\max\{\beta^2\cdot\vD^{-1}_\beta(z),(1-\beta)\cdot[\vD^{-1}_\beta(t)+\vD^{-1}_\beta(x_2)]\}\\
    &=\arg\min_{\beta\in [0,1]}\max\{\beta^2,k(1-\beta)\}\\
    &=\sqrt{k^2+2k}-k.\\
\end{align*}

\end{proof}

\section{ Different Attacks under RDGA setting}
\label{sec:appendix-diff-attacks}
In Section~\ref{sec:directed-attack}, we mention that the greedy approaches such as FGA~\cite{mujkanovic2022defenses} and Nettack~\citep{zugner2018adversarial} tend to be relatively weaker and  we adopt PGD~\citep{xu2019topology} under RDGA setting in our work. Here, we also provide the experimental results of FGA~\cite{mujkanovic2022defenses} and Nettack~\citep{zugner2018adversarial} under RDGA setting to validate the aforementioned statement. We present the results on Cora-ML of 50\% and 100\% budgets in Table~\ref{tab:diff-attack-50} and Table~\ref{tab:diff-attack-100}.
\begin{table}[!ht]
    \centering
    \caption{Classification accuracy (\%) under different attacks (Cora-ML, budget=50\%)}
    \renewcommand\arraystretch{1.0}
    \label{tab:diff-attack-50}

    \resizebox{0.7\linewidth}{!}{%
    \begin{tabular}{ccccc}
    \toprule
     Attack $\backslash$ Model&BBRW-GCN&BBRW-APPNP&BBRW-SoftMedian\\ 
    \hline
         RDGA-FGA&  84.5±6.5	&85.0±7.1&	91.5±5.9 \\        
          RDGA-Nettack&   88.0±6.0	&84.5±6.1&	89.0±5.4 \\ 
           RDGA-PGD&   85.0±6.3&	83.0±6.4	&88.0±5.1 \\ 
    \bottomrule
    \end{tabular}
    }
\end{table}

\begin{table}[!ht]
    \centering
    \caption{Classification accuracy (\%) under different attacks (Cora-ML, budget=100\%)}
    \renewcommand\arraystretch{1.3}
    \label{tab:diff-attack-100}

    \resizebox{0.7\linewidth}{!}{%
    \begin{tabular}{ccccc}
    \toprule
     Attack $\backslash$ Model&BBRW-GCN&BBRW-APPNP&BBRW-SoftMedian\\ 
    \hline
         RDGA-FGA&  77.0±10.1&	77.0±6.4&	92.0±6.4 \\        
          RDGA-Nettack&   85.5±7.2&	79.0±6.6&	86.5±6.7\\ 
           RDGA-PGD&   75.0±10.2	&69.0±9.7	&84.5±8.8\\ 
    \bottomrule
    \end{tabular}
    }
\end{table}

\newpage

\section{Supplementary Experiments}
\label{sec:appendix-supp-figure}



\subsection{More Results with Low Budgets}
\label{sec:results_low_budget}

\begin{table}[!ht]
    \centering
    \caption{ Classification accuracy (\%) under different perturbation rates of graph attack. 
    The best results are in \textbf{bold}, and the second-best results are \underline{underlined}. (Cora-ML) }
    \vspace{-0.1in}
    \renewcommand\arraystretch{1.3}
    \label{tab:acc-cora-low-budget}
    \resizebox{0.9\linewidth}{!}{%
{
    \begin{tabular}{ccccccccccccccc}
    \toprule
     \multirow{2}{*}{Method}&\multicolumn{2}{c}{5\%}&\multicolumn{2}{c}{10\%}&\multicolumn{2}{c}{15\%}&\multicolumn{2}{c}{20\%}\\
    &Transfer&Adaptive&Transfer&Adaptive&Transfer&Adaptive& Transfer&Adaptive \\ \hline

MLP
& 73.5$\pm$7.4 & 73.5$\pm$7.4
& 73.5$\pm$7.4 & 73.5$\pm$7.4
& 73.5$\pm$7.4 & 73.5$\pm$7.4
& 73.5$\pm$7.4 & 73.5$\pm$7.4 \\
\hline
DGCN
& 87.0$\pm$8.7 & $\backslash$
& 84.5$\pm$9.8 & $\backslash$
& 81.5$\pm$10.8 & $\backslash$
& 79.0$\pm$11.9 & $\backslash$ \\

DiGCN
& 78.0$\pm$7.3 & $\backslash$
& 71.0$\pm$7.1 & $\backslash$
& 64.0$\pm$7.0 & $\backslash$
& 57.0$\pm$6.8 & $\backslash$ \\

MagNet
& 88.0$\pm$3.3 & $\backslash$
& 86.5$\pm$3.9 & $\backslash$
& 85.0$\pm$5.5 & $\backslash$
& 78.5$\pm$8.1 & $\backslash$ \\
\hline
Jaccard-GCN
& 86.5$\pm$6.8 & 82.5$\pm$6.8
& 82.0$\pm$7.1 & 78.0$\pm$7.1
& 78.0$\pm$7.3 & 74.0$\pm$7.3
& 73.5$\pm$7.6 & 69.5$\pm$7.6 \\

RGCN
& 87.5$\pm$6.7 & 87.0$\pm$6.8
& 87.0$\pm$6.4 & 86.5$\pm$6.8
& 85.0$\pm$6.7 & 84.5$\pm$6.9
& 81.0$\pm$8.6 & 77.5$\pm$10.3 \\

GRAND
& 83.0$\pm$6.3 & 74.0$\pm$6.6
& 81.0$\pm$6.5 & 72.0$\pm$6.8
& 78.5$\pm$6.6 & 69.5$\pm$6.9
& 76.5$\pm$6.8 & 67.5$\pm$7.1 \\

GNNGuard
& 89.5$\pm$5.3 & 77.0$\pm$5.7
& 89.0$\pm$5.6 & 76.5$\pm$6.5
& 88.5$\pm$5.8 & 76.0$\pm$7.2
& 88.0$\pm$6.1 & 75.5$\pm$8.0 \\

GARNET
& 89.0$\pm$5.8 & 86.0$\pm$6.7
& 87.0$\pm$3.4 & 84.0$\pm$6.0
& 85.5$\pm$8.3 & 82.5$\pm$6.2
& 84.0$\pm$5.6 & 81.0$\pm$6.5 \\

HANG
& 89.5$\pm$6.1 & 86.5$\pm$7.0
& 87.5$\pm$6.6 & 84.5$\pm$7.2
& 86.0$\pm$7.4 & 83.0$\pm$9.5
& 84.5$\pm$8.7 & 81.5$\pm$6.3 \\

NoisyGNN
& 88.0$\pm$7.5 & 85.0$\pm$5.9
& 86.0$\pm$4.7 & 83.0$\pm$5.8
& 84.5$\pm$4.0 & 81.5$\pm$6.6
& 83.0$\pm$6.7 & 80.0$\pm$9.3 \\

RUNG
& \underline{91.5$\pm$5.6} & 88.5$\pm$8.5
& 90.5$\pm$6.7 & 87.5$\pm$5.8
& 89.0$\pm$7.0 & 86.0$\pm$6.1
& 88.5$\pm$8.3 & 85.0$\pm$6.4 \\

\hline
GCN
& 88.0$\pm$7.8 & 87.5$\pm$6.3
& 87.5$\pm$5.8 & 87.0$\pm$6.9
& 83.0$\pm$5.1 & 82.5$\pm$7.7
& 76.5$\pm$3.8 & 76.0$\pm$9.6 \\

BBRW-GCN
& 90.0$\pm$6.5 & \underline{89.5$\pm$5.9}
& \underline{90.0$\pm$4.5} & \underline{89.5$\pm$5.1}
& 89.5$\pm$4.2 & \underline{89.0$\pm$6.3}
& \underline{89.5$\pm$5.7} & \underline{89.0$\pm$7.6} \\
\hline
APPNP
& 88.5$\pm$5.7 & 87.5$\pm$5.8
& 87.0$\pm$6.6 & 86.0$\pm$7.0
& 85.0$\pm$7.6 & 84.0$\pm$8.1
& 83.5$\pm$8.5 & 82.5$\pm$9.3 \\

BBRW-APPNP
& 90.5$\pm$6.0 & 89.0$\pm$5.3
& \underline{90.0$\pm$5.1} & 88.5$\pm$5.2
& \underline{90.0$\pm$5.2} & 88.5$\pm$5.3
& \underline{89.5$\pm$5.3} & 88.0$\pm$5.5 \\
\hline
SoftMedian
& 90.5$\pm$7.8 & 87.5$\pm$5.2
& 89.5$\pm$4.1 & 86.5$\pm$6.6
& 88.0$\pm$3.4 & 85.0$\pm$6.6
& 87.0$\pm$7.7 & 84.0$\pm$6.3 \\

BBRW-SoftMedian
& \textbf{92.0$\pm$3.7} & \textbf{92.0$\pm$4.2}
& \textbf{92.0$\pm$5.8} & \textbf{92.0$\pm$4.6}
& \textbf{91.5$\pm$7.9} & \textbf{92.0$\pm$4.8}
& \textbf{91.5$\pm$6.0} & \textbf{92.0$\pm$5.8}  \\
       
        \bottomrule
    \end{tabular}
}
}
\end{table}

\begin{table}[!ht]
    \centering
    \caption{ 
    Classification accuracy (\%) under different perturbation rates of graph attack. 
    The best results are in \textbf{bold}, and the second-best results are \underline{underlined}. (Citeseer)}
    \vspace{-0.1in}
    \renewcommand\arraystretch{1.3}
    \label{tab:acc-citeseer-low-budget}
    \resizebox{0.9\linewidth}{!}{%

    \begin{tabular}{ccccccccccccccc}
    \toprule
     \multirow{2}{*}{Method}&\multicolumn{2}{c}{5\%}&\multicolumn{2}{c}{10\%}&\multicolumn{2}{c}{15\%}&\multicolumn{2}{c}{20\%}\\
    &Transfer&Adaptive&Transfer&Adaptive&Transfer&Adaptive& Transfer&Adaptive \\ \hline
MLP & 49.0$\pm$9.4 & 49.0$\pm$9.4 & 49.0$\pm$9.4 & 49.0$\pm$9.4 & 49.0$\pm$9.4 & 49.0$\pm$9.4 & 49.0$\pm$9.4 & 49.0$\pm$9.4 \\
\midrule
DGCN   & 62.0$\pm$7.3 & \textbackslash & 60.0$\pm$7.5 & \textbackslash & 58.0$\pm$7.8 & \textbackslash & 56.0$\pm$8.0 & \textbackslash \\
DiGCN  & 61.0$\pm$9.0 & \textbackslash & 56.0$\pm$9.4 & \textbackslash & 51.5$\pm$9.7 & \textbackslash & 46.5$\pm$10.1 & \textbackslash \\
MagNet & 64.5$\pm$7.0 & \textbackslash & 61.5$\pm$8.1 & \textbackslash & 58.0$\pm$9.1 & \textbackslash & 55.0$\pm$10.2 & \textbackslash \\
\midrule
Jaccard-GCN & 55.0$\pm$7.3 & 53.5$\pm$7.6 & 52.5$\pm$7.4 & 50.0$\pm$8.1 & 50.0$\pm$7.6 & 46.0$\pm$8.6 & 48.0$\pm$7.7 & 42.0$\pm$9.1 \\
RGCN        & 56.0$\pm$7.5 & 56.0$\pm$7.7 & 50.5$\pm$7.9 & 50.5$\pm$8.3 & 45.5$\pm$8.3 & 45.0$\pm$8.9 & 40.0$\pm$8.7 & 39.5$\pm$9.5 \\
GRAND       & \underline{65.5$\pm$6.1} & 65.0$\pm$6.6 & 63.0$\pm$6.1 & 63.0$\pm$7.2 & \underline{61.0$\pm$6.2} & \underline{60.5$\pm$7.7} & 58.5$\pm$6.2 & 58.0$\pm$8.3 \\
GNNGuard    & 58.5$\pm$7.5 & 57.0$\pm$7.6 & 56.5$\pm$7.8 & 53.5$\pm$7.9 & 54.0$\pm$8.1 & 50.0$\pm$8.3 & 52.0$\pm$8.4 & 46.5$\pm$8.6 \\
GARNET      & 60.0$\pm$5.2 & 59.5$\pm$7.7 & 58.0$\pm$6.4 & 57.0$\pm$7.2 & 56.0$\pm$8.6 & 54.5$\pm$7.2 & 54.5$\pm$4.8 & 52.5$\pm$7.6 \\
HANG        & 61.0$\pm$5.6 & 60.5$\pm$7.4 & 59.0$\pm$6.7 & 58.5$\pm$7.8 & 57.0$\pm$6.8 & 56.0$\pm$7.2 & 55.0$\pm$5.9 & 53.5$\pm$7.4 \\
NoisyGNN    & 59.0$\pm$8.7 & 58.5$\pm$6.8 & 57.0$\pm$4.9 & 56.0$\pm$6.7 & 55.0$\pm$6.1 & 54.0$\pm$7.2 & 53.0$\pm$4.3 & 51.5$\pm$7.4 \\
RUNG        & 63.0$\pm$5.8 & 62.5$\pm$6.1 & 61.5$\pm$6.1 & 60.5$\pm$7.3 & 60.5$\pm$8.4 & 59.0$\pm$7.3 & 59.0$\pm$4.7 & 57.0$\pm$7.6 \\
\midrule
GCN & 54.5$\pm$7.2 & 54.5$\pm$6.4 & 50.0$\pm$6.0 & 50.0$\pm$7.2 & 45.5$\pm$4.9 & 45.5$\pm$7.3 & 41.0$\pm$6.7 & 41.0$\pm$8.4 \\
BBRW-GCN (Ours) & 59.0$\pm$7.5 & 57.5$\pm$8.0 & 57.0$\pm$7.5 & 54.0$\pm$8.6 & 54.5$\pm$7.6 & 49.5$\pm$9.1 & 52.5$\pm$7.6 & 44.5$\pm$9.7 \\
\midrule
APPNP & \textbf{68.5$\pm$6.7} & \textbf{67.5$\pm$6.0} & \underline{64.5$\pm$7.4} & \underline{63.5$\pm$6.1} & \underline{61.0$\pm$8.1} & 59.5$\pm$6.1 & 57.0$\pm$8.8 & 55.0$\pm$6.2 \\
BBRW-APPNP (Ours) & \textbf{68.5$\pm$5.2} & \underline{67.0$\pm$5.5} & \textbf{67.5$\pm$6.0} & \textbf{65.0$\pm$6.5} & \textbf{67.0$\pm$6.8} & \textbf{62.5$\pm$7.0} & \textbf{66.5$\pm$7.5} & \textbf{60.5$\pm$7.6} \\
\midrule
SoftMedian & 60.5$\pm$7.4 & 60.5$\pm$6.7 & 59.5$\pm$4.9 & 59.5$\pm$6.3 & 58.0$\pm$5.3 & 58.0$\pm$7.6 & 57.0$\pm$8.8 & 57.0$\pm$7.3 \\
BBRW-SoftMedian (Ours) & 59.5$\pm$5.3 & 59.5$\pm$7.4 & 59.5$\pm$6.4 & 59.5$\pm$7.7 & 59.5$\pm$2.6 & 59.5$\pm$7.4 & \underline{59.0$\pm$6.7} & \underline{59.0$\pm$7.9} \\

        \bottomrule
    \end{tabular}
}
\end{table}


\newpage




\subsection{Ablation Study on Masking Rate of Transfer Attack}

\begin{wrapfigure}{r}{0.44\textwidth}
\vspace{-0.2in}
\begin{center}
\includegraphics[width=0.40\columnwidth]{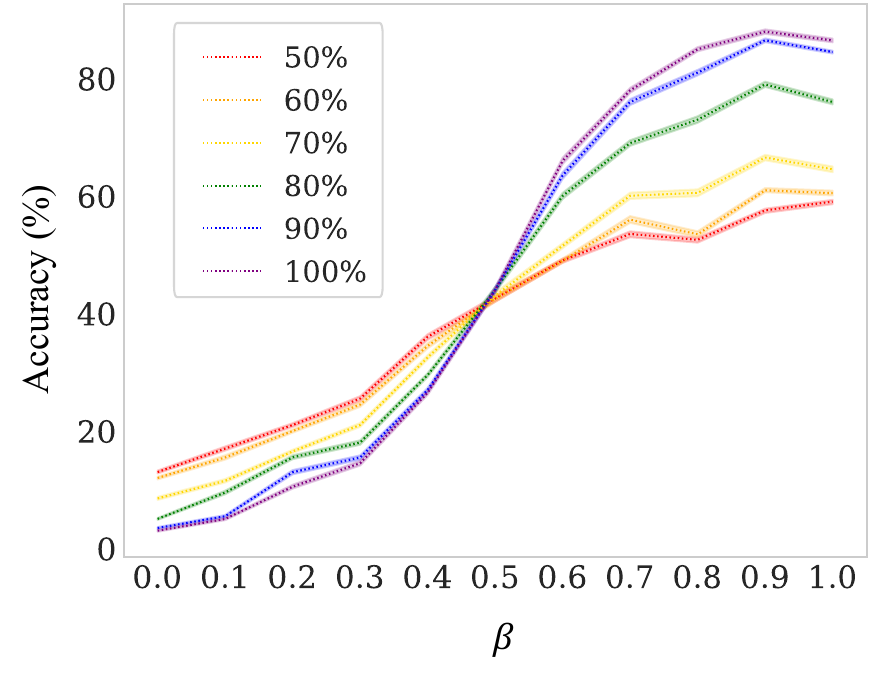}
\end{center}
\vspace{-0.2in}
\caption{Masking rates under transfer attack.}
 \vspace{-0.2in}
\label{fig:masking_rate_transfer}
\vspace{-0.1in}
\end{wrapfigure}

In addition to the ablation study on masking rate of adaptive attack, we also provide the ablation experiment on the transfer attack in 
Fig.~\ref{fig:masking_rate_transfer}
. The $\beta$ in the x-axis denotes the weight in the BBRW module, we vary the masking rate from 50\% to 100\%. As can be seen in the figure, when the masking rate is larger, we need a more biased weight to defend the in-links attacks. Particularly, when $\beta=0.5$, the BBRW is equivalent to the undirected random walk propagation. Therefore, the performance of BBRW ($\beta=0.5$) is insensitive to the masking rate cause undirected GNNs can not differentiate the direction of links.

\newpage

\subsection{Ablation Study on Attacking Patterns.}
\begin{figure}[h!]
 \centering
 \includegraphics[width=0.75\textwidth]{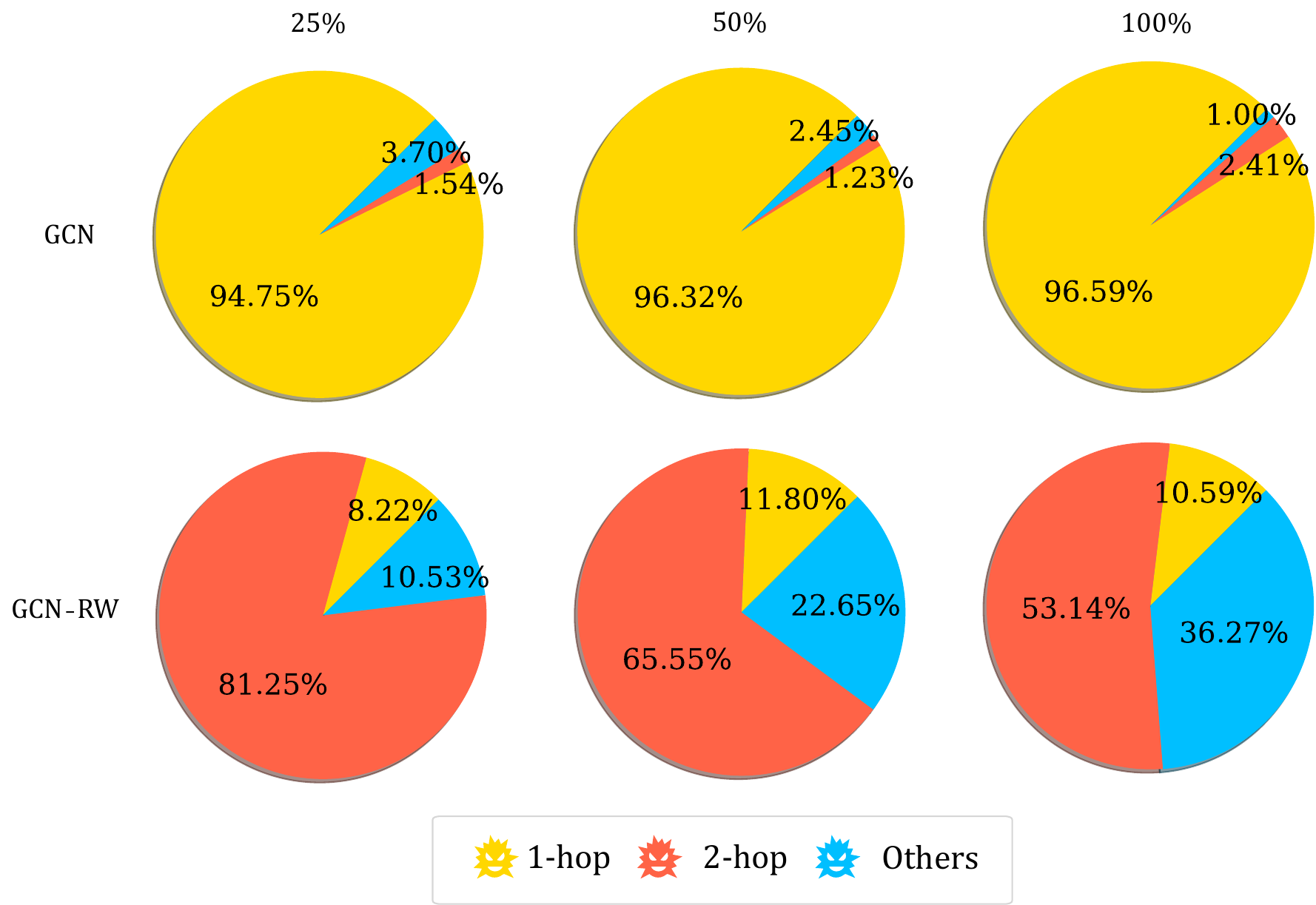}
 \caption{Distributions of adversarial links against different victim models (GCN and GCN-RW) under different attack budgets (25\%, 50\%, 100\%). The yellow portion represents attacks by 1-hop neighbors on the target (direct in-link attacks); the red portion represents attacks by 2-hop neighbors on the targets’ neighbors (indirect out-link attacks); and the blue portion represents other attacks.}
 \label{fig:pie-rw-all}
\end{figure}

Figure~\ref{fig:pie-rw-all} presents the distribution of adversarial links under different attack budgets, indicating the adversary's distinct attacking behaviors when attacking different victim models adaptively.

\end{document}